\documentclass{article} % For LaTeX2e
\usepackage{iclr2027_conference,times}

\usepackage{amsmath,amsfonts,bm}

\def\eqref#1{equation~\ref{#1}}
\def\1{\bm{1}}

\DeclareMathAlphabet{\mathsfit}{\encodingdefault}{\sfdefault}{m}{sl}
\SetMathAlphabet{\mathsfit}{bold}{\encodingdefault}{\sfdefault}{bx}{n}

\usepackage{hyperref}
\usepackage{xcolor, algorithm, algpseudocode}
\definecolor{onpol}{RGB}{0,90,160}
\definecolor{delop}{RGB}{210,120,55}
\newcommand{\op}[1]{\textcolor{onpol}{#1}}
\newcommand{\delop}[1]{\textcolor{delop}{#1}}

\usepackage{url}
\usepackage{amsthm, amsmath, amssymb}
\usepackage{graphicx}
\usepackage{subcaption}
\usepackage{multirow, booktabs, arydshln}
\usepackage{pgfplots}
\usepgfplotslibrary{groupplots}
\pgfplotsset{compat=1.18}
\newcommand{\ffhqfile}[1]{a_star_weights_1000_100_#1.csv}
\newcommand{\imagenetfile}[1]{a_star_weights_1000_50_imagenet_#1.csv}
\newcommand{\afhqfile}[1]{afhq_cat_flow_indp_#1.csv}
\newcommand{\celebafile}[1]{celeba_flow_indp_#1.csv}

\title{Getting Your Guidance Weights Right in diffusion and flow-matching posterior sampling}

\author{L. Moroy, \\
\small Heriot-Watt University, EPS\\
Edinburgh, UK \\
\And
J.-F. Giovannelli \\
\small IMS (Univ. Bordeaux, CNRS, BINP) \\
Bordeaux, France \\
\And
Y. Altmann \\
\small Heriot-Watt University, EPS\\
Edinburgh, UK \\
\AND
S. McLaughlin \\
\small Heriot-Watt University, EPS\\
Edinburgh, UK \\
\And
F. Champagnat \\
\small ONERA\\
Paris, France \\
\And
G. Bourmaud \\
\small IMS (Univ. Bordeaux, CNRS, BINP) \\
Bordeaux, France \\
}

\iclrfinalcopy % Uncomment for camera-ready version, but NOT for submission.
\begin{document}

\maketitle

\begin{abstract}
Training-free posterior sampling methods, also known as Plug-and-Play methods, leverage pretrained unconditional diffusion or flow-matching models to solve inverse problems. Most existing approaches rely on guidance weights to balance, at each time step, prior information from the unconditional score or velocity network with measurement consistency, yet the tuning of these weights is often not discussed and is largely left to heuristics.
We introduce a simple and principled offline strategy for automatically tuning these guidance weights. Our key observation is that, at each time step, the conditional denoising score-matching objective for diffusion models, or the conditional flow-matching objective for flow-matching models, is a least-squares objective. Therefore, when the conditional prediction is expressed as a weighted sum of the unconditional network output and a measurement-guidance term, optimizing over these weights reduces to a two-dimensional linear least-squares problem.
The resulting time-dependent guidance weights can be optimized offline for a given measurement operator, noise level and sampler at the cost of a single minibatch of sampling trajectories, 
without retraining or fine-tuning the pretrained generative model.
Instantiated with the standard Tweedie-based measurement-consistency term, our approach improves posterior sampling and achieves state-of-the-art reconstruction performance across diffusion- and flow-matching-based methods. Moreover, the optimized guidance weights enable diffusion samplers to reduce the number of sampling steps from 1000 to 50 with no significant degradation in reconstruction quality. 
Code  will be made available. 
\end{abstract}

\section{Introduction}
Diffusion models~\citep{ho2020denoising, karras2022elucidating, song2019generative, song2020score, song2020improved} and flow matching models~\citep{lipman2022flow, liu2022rectified, tong2023improving, albergo2025stochastic, liu2022flow} have emerged as dominant families of iterative generative models, achieving remarkable success in synthesizing high-fidelity samples across a wide range of data modalities. However, training these models remains computationally expensive and can require substantial computational resources and time. In inverse problems, where the exact measurement operator and noise characteristics often depend on the acquisition setting specified at inference time, retraining or fine-tuning a generative model for each new acquisition setting is often impractical. Practitioners instead commonly rely on {training-free} methods~\citep{chung2022diffusion, song2023pseudoinverse, boys2023tweedie, kim2025flowdps, martin2025pnp}, also referred to as Plug-and-Play methods. These methods leverage {unconditionally pretrained} generative models to perform approximate posterior sampling without requiring additional training for different measurement operators or noise levels.

Most existing training-free methods~\citep{chung2022diffusion, song2023pseudoinverse, boys2023tweedie, kim2025flowdps, martin2025pnp} incorporate a measurement-consistency step at each time step to modify the unconditional sampling direction, thereby steering the sampling trajectory toward samples consistent with the  measurement. Repeating this procedure throughout the sampling process yields a sample from the approximate posterior distribution. Such methods usually rely on guidance weights to balance, at each time step, the prior information provided by the unconditional score or velocity network against measurement consistency. However, the tuning of these weights is often not discussed and is largely left to heuristics.

In practice, the measurement setup may frequently change, for instance by modifying the number or locations of sensors, or their internal acquisition settings~\citep{donoho2006compressed}. Such modifications require the guidance weights to be re-tuned, introducing an additional and time-consuming hyperparameter optimization step. Consequently, the need to tune guidance weights represents a key practical bottleneck of existing training-free posterior sampling methods.

This motivates the development of a strategy to automatically and efficiently determine appropriate guidance weights for a given measurement setup. Importantly, such a strategy must introduce negligible computational overhead. From a practitioner's perspective, guidance selection should take only a few seconds, ruling out iterative optimizers (\emph{e.g.} gradient descent, etc.) that require repeated ODE/SDE integrations.
In this paper, we introduce a simple and principled offline calibration strategy for automatically tuning time-dependent guidance weights. Our contributions are as follows:
\begin{itemize}
    \item  We make the key observation that, at each time step, the conditional denoising score-matching objective for diffusion models and the conditional flow-matching objective for flow-matching models are least-squares objectives. Therefore, when the conditional prediction is expressed as a weighted sum of the unconditional network output and a measurement-guidance term, optimizing these weights reduces to a two-dimensional linear least-squares problem. This yields a highly efficient, simulation-free strategy for calibrating time-dependent guidance weights.

    \item While theoretically grounded, the simulation-free calibration strategy does not guarantee improved posterior sampling trajectories, since neither the measurement-consistency term nor the unconditional network is necessarily optimal. To address this limitation, we propose a greedy SIMulation-based CAlibration strategy (SIMCA), which sequentially calibrates the guidance weight at each time step while keeping the weights calibrated at previous time steps fixed. SIMCA requires only a single minibatch of sampling trajectories, making it practical for tuning time-dependent guidance weights in a given measurement setup.
    
    \item We empirically demonstrate that SIMCA, instantiated with the standard Tweedie-based measurement-consistency term, matches or surpasses state-of-the-art training-free methods across multiple inverse problems and datasets, using both diffusion and flow-matching backbones. 
    
    \item We further show that SIMCA enables diffusion samplers to reduce the number of sampling steps from 1000 to 50, a $\times 20$ reduction in inference cost, without significant degradation in reconstruction quality.
\end{itemize}

\section{Background on training-free posterior sampling}%{Background of training-free generative models for solving inverse problems}
\label{sec:background}

\paragraph{Unconditional diffusion models and (Gaussian) flow-matching.}%Generative models along Gaussian paths.}
Let $p_0$ denote the data distribution on $\mathbb{R}^d$. Diffusion~\citep{ho2020denoising, karras2022elucidating, song2019generative, song2020score, song2020improved} and (Gaussian) flow matching\footnote{For flow-matching, we focus here on the independent coupling case, as it allows us to establish the equivalence between diffusion models and Gaussian flow matching~\citep{gao2025diffusionmeetsflow}. In our experiments, however, we evaluate velocity networks trained using both independent and OT couplings.
} \cite{lipman2022flow, liu2022rectified, albergo2025stochastic} models both generate samples from $p_0$ by transporting Gaussian noise along a prescribed family of marginals, obtained by interpolating data and noise,
\begin{equation}
    \mathbf{x}_t = \alpha_t \mathbf{x}_0 + \sigma_t \boldsymbol{\epsilon},
    \qquad \mathbf{x}_0 \sim p_0, \quad \boldsymbol{\epsilon} \sim \mathcal{N}(\mathbf{0},\mathbf{I}),
    \qquad t \in [0,1],
    \label{eq:path}
\end{equation}
so that $p_t(\mathbf{x}_t \mid \mathbf{x}_0) = \mathcal{N}(\alpha_t \mathbf{x}_0, \sigma_t^2\mathbf{I})$. Throughout, $t=0$ denotes data and $t=1$ noise. The schedule $(\alpha_t,\sigma_t)$ is chosen with $\alpha_0 \approx 1$, $\sigma_0 \approx 0$, $\alpha_1 \approx 0$ and $\sigma_1 \approx 1$, so that $p_1 \approx \mathcal{N}(\mathbf{0},\mathbf{I})$ and sampling can be initialized from pure noise. The variance-preserving diffusion schedule $\alpha_t^2 + \sigma_t^2 = 1$ and the optimal-transport interpolant $\alpha_t = 1-t$, $\sigma_t = t$ are two instances~\citep{song2020score, lipman2022flow}.

The two families differ in the vector field they learn. % along this shared path: 
Diffusion models regress the score \linebreak $\mathbf{s}_{\boldsymbol{\theta}}(\mathbf{x}_t,t) \approx \nabla_{\mathbf{x}_t}\log p_t(\mathbf{x}_t)$ by (often weighted) Denoising Score Matching (DSM)~\citep{vincent2011connection, song2019generative, song2020score}
\begin{equation}
    \mathcal{L}_{\text{DSM}}(\boldsymbol{\theta})= \mathbb{E}\left[ \left\Vert \mathbf{s}_{\boldsymbol{\theta}}(\mathbf{x}_t,t) +\frac{\boldsymbol{\epsilon}}{\sigma_t} \,\right\Vert^2 \right], 
    \label{eq:DSM}
\end{equation}
where $\left\Vert \cdot \right\Vert$ is the L2-norm, while flow matching models regress the velocity $\mathbf{v}_{\boldsymbol{\theta}}(\mathbf{x}_t,t)$ %of the probability-flow ODE 
by Conditional Flow Matching (CFM)~\citep{lipman2022flow, liu2022rectified}
\begin{equation}
    \mathcal{L}_{\text{CFM}}(\boldsymbol{\theta})= \mathbb{E}\left[ \left\Vert \mathbf{v}_{\boldsymbol{\theta}}(\mathbf{x}_t,t) -\dot{\alpha}_t \mathbf{x}_0 - \dot{\sigma}_t \boldsymbol{\epsilon} \,\right\Vert^2 \right], 
    \label{eq:CFM}
\end{equation}
where $\dot{\alpha}_t$ and $\dot{\sigma}_t$ denote the time derivatives. In the rest of the paper, we employ unified notations as our calibration strategy can be applied in both cases:
\begin{equation}
    \mathcal{L}(\boldsymbol{\theta})= \mathbb{E}\left[ \left\Vert \mathbf{u}_{\boldsymbol{\theta}}(\mathbf{x}_t,t) - \boldsymbol{\psi}_t(\mathbf{x}_0,\boldsymbol{\epsilon}) \,\right\Vert^2 \right], 
    \label{eq:loss_unif}
\end{equation}
where \begin{equation}
    \boldsymbol{\psi}_t(\mathbf{x}_0,\boldsymbol{\epsilon}) =
    \begin{cases}
        -\boldsymbol{\epsilon}/\sigma_t & \text{(diffusion case)},\\[2pt]
        \dot{\alpha}_t \mathbf{x}_0 + \dot{\sigma}_t \boldsymbol{\epsilon} & \text{(flow-matching case)}.
    \end{cases}
    \label{eq:psi}
\end{equation}

Sampling integrates the corresponding dynamics backward on a grid $1 = t_K > \dots > t_0 = 0$, where the number of steps $K$ trades sample quality against computational cost. In this paper, we consider both deterministic and stochastic samplers, namely DDPM~\citep{ho2020denoising}, Euler-Maruyama on the reverse SDE, and the Euler step of the flow ODE:
\begin{equation}
    \mathbf{x}_{t_{k-1}} = \Phi\left(\mathbf{x}_{t_k},\mathbf{u}_{\boldsymbol{\theta}}(\mathbf{x}_{t_k},{t_k}),{t_k}\right).
    \label{eq:sampler}
\end{equation}
%where
%\begin{equation}
%\Phi\left(\mathbf{x},\mathbf{u},t\right) = \mathbf{m}(\mathbf{x},{t}) + c({t})\,\mathbf{u} + s({t})\,\mathbf{z},
%    \qquad \mathbf{z}\sim\mathcal{N}(\mathbf{0},\mathbf{I})
%\end{equation}
%\begin{equation}
%    \mathbf{x}_{t-\Delta t} = \mathbf{m}(\mathbf{x}_t,t) + c(t)\,\mathbf{u}_{\boldsymbol{\theta}}(\mathbf{x}_t,t) + s(t)\,\mathbf{z},
%    \qquad \mathbf{z}\sim\mathcal{N}(\mathbf{0},\mathbf{I}),
%    \label{eq:sampler}
%\end{equation}

%with $\mathbf{m}$ affine in $\mathbf{x}$ and $c,s$ deterministic scalars.%, $s 

%\equiv 0$ recovering deterministic samplers; the coefficients for each are collected in Appendix~\ref{app:samplers}. %Because the field enters Eq.~\eqref{eq:sampler} linearly, rescaling it acts affinely on the next iterate, which is the structure our calibration procedure exploits in Section~\ref{sec:method}.

\paragraph{Inverse problems and Training-free posterior sampling.}
We seek to recover an unknown signal $\mathbf{x}_0$ from a degraded observation
\begin{equation}
    \mathbf{y} = \mathcal{A}\mathbf{x}_0 + \mathbf{n},
    \qquad \mathbf{n}\sim\mathcal{N}(\mathbf{0},\sigma_m^2\mathbf{I}),
    \label{eq:obs}
\end{equation}
where $\mathcal{A}$ is a known linear measurement operator, the setting we consider throughout. %Such problems are ill posed, since many signals explain the same observation, which motivates sampling the posterior distribution $p(\mathbf{x}_0\mid\mathbf{y}) \propto p(\mathbf{x}_0)\,p(\mathbf{y}\mid\mathbf{x}_0)$ rather than returning a point estimate.

%Applying Eq.~\ref{eq:path} to $p(\mathbf{x}_0 \mid \mathbf{y})$ defines conditional marginals $p_t(\mathbf{x}_t\mid\mathbf{y})$, so p 
Posterior sampling amounts to running the same dynamics driven by a learned conditional field $\mathbf{c}_{\boldsymbol{\theta}}(\mathbf{x}_t,\mathbf{y},t)$
 instead of the unconditional one $\mathbf{u}_{\boldsymbol{\theta}}(\mathbf{x}_t,t)$. Training-free methods aim to approximate that conditional field $\mathbf{c}_{\boldsymbol{\theta}}(\mathbf{x}_t,\mathbf{y},t)$ leveraging an unconditionally pretrained network $\mathbf{u}_{\boldsymbol{\theta}}(\mathbf{x}_t,t)$, along with a measurement-consistency mechanism,
without requiring additional training for different measurement operators $\mathcal{A}$ or noise levels $\sigma_m$~\citep{song2020score, chung2022diffusion, song2023pseudoinverse, boys2023tweedie}.

\paragraph{Measurement-consistency mechanism.}%Injecting measurement information.}
Many different measurement-consistency mechanisms have been proposed for both diffusion and flow-matching models. Existing strategies include Tweedie-based approximations~\citep{chung2022diffusion, song2023pseudoinverse, boys2023tweedie}, spectral methods exploiting the singular value decomposition of $\mathcal{A}$~\citep{kawar2022denoising, kawar2022jpeg}, sequential Monte Carlo~\citep{cardoso2024monte, dou2024diffusion, kelvinius2025solving}, conversion of the velocity field to a score followed by diffusion-style guidance~\citep{kim2025flowdps, song2026flow}, and schemes that interleave generative steps with optimization~\citep{song2021solving, graikos2022diffusion, mardani2024variational} or projection~\citep{wang2022zero, chung2022improving, song2023pseudoinverse} steps.
In this paper, we focus on a specific measurement-consistency mechanism that approximates the conditional field 
with a weighted sum of the unconditional network output and
a measurement-guidance term $\mathbf{f}(\mathbf{x}_t,\mathbf{y},t)$:
\begin{equation}
    \mathbf{c}_{\boldsymbol{\theta}}(\mathbf{x}_t,\mathbf{y},t) \;\approx\; w_{u,t}\,\mathbf{u}_{\boldsymbol{\theta}}(\mathbf{x}_t,t) + w_{f,t}\,\mathbf{f}(\mathbf{x}_t,\mathbf{y},t),
    \label{eq:additive}
\end{equation}
where $\mathbf{w}_{t}=\left[w_{u,t},w_{f,t}\right]^\top$ are the scalar guidance weights. A well-known example of an additive measurement-consistency term is the Tweedie-based measurement-consistency term introduced in~\citep{chung2022diffusion}, which we employ in our experiments:
\begin{equation}
    \mathbf{f}(\mathbf{x}_t,\mathbf{y},t) \propto \nabla_{\mathbf{x}_t} \left\Vert \mathbf{y} - \mathcal{A}\hat{\mathbf{x}}_0(\mathbf{x}_t) \right\Vert^2.
    \label{eq:DPS_meas_cons}
\end{equation}
%\begin{equation}
%    \mathbf{f}(\mathbf{x}_t,\mathbf{y},t) \propto \mathcal{J}_{\hat{\mathbf{x}}_0}(\mathbf{x}_t)^\top
%    \mathcal{A}^\top
%    \left(\mathbf{y} - \mathcal{A}\hat{\mathbf{x}}_0(\mathbf{x}_t)\right),
%    \label{eq:DPS_meas_cons}
%\end{equation}
%
%
%where $\mathcal{A}^\top$ is the adjoint operator of $\mathcal{A}$,  
where $\hat{\mathbf{x}}_0(\mathbf{x}_t)$ is the Tweedie estimate~\citep{efron2011tweedie}.
%, and $\mathcal{J}_{\hat{\mathbf{x}}_0}(\mathbf{x}_t)$ is its Jacobian matrix. 
This paper focuses on the tuning of these guidance weights, that are often not discussed and largely left to heuristics.

\section{Related Work}
\paragraph{Guidance weights in training-free posterior sampling.}
Additive training-free samplers weight their measurement-consistency term either heuristically, as the residual-normalized step of DPS~\citep{chung2022diffusion} and the hand-set schedules of flow-based samplers~\citep{kim2025flowdps, martin2025pnp}, or through a Gaussian model of the denoising posterior, as in $\Pi$GDM~\citep{song2023pseudoinverse}, TMPD~\citep{boys2023tweedie}, their refinements~\citep{peng2024improving}, and their flow counterpart OT-ODE~\citep{pokle2024training}. The latter derive the weight from an assumed posterior covariance rather than tuning it, and thereby improve both the measurement term and its effective scale; our ablations suggest that this second effect accounts for much of their benefit (Section~\ref{sec:ablations}). What none of them does is fit the weight to a criterion measurable from data, and all fix the weight of the pretrained field to one.

\paragraph{Adaptive guidance.}
Time-varying weights have been studied mainly for classifier-free guidance~\citep{malarz2025classifier}, where the weight trades sample quality against conditioning fidelity and, having no reference value, is set heuristically. In training-free inverse problems the guided field approximates a well-defined conditional field, so its approximation error can be measured against the tractable target of Eq.~\ref{eq:psi} and minimized in closed form rather than tuned. SAIP~\citep{wang2026saip} also adapts the balance between the two terms, but through a single coupled parameter, estimated online and stated against the intractable posterior score. Methodologically, our closed-form solve from Monte Carlo moments follows Analytic-DPM~\citep{bao2022analytic}, applied there to the unconditional reverse variance.

\section{Method}
\label{sec:method}
Given a pretrained unconditional score or velocity network $\mathbf{u}_{\boldsymbol{\theta}}$ and a user-specified measurement-consistency term $\mathbf{f}$, we seek to automatically and efficiently calibrate time-dependent
guidance weights $\mathbf{w} = \left\{\mathbf{w}_{t_k}\right\}_{k=1\ldots K}$ for a given measurement operator $\mathcal{A}$ and noise level $\sigma_m$. Importantly, the proposed strategy must introduce negligible computational overhead, since from a practitioner's perspective, calibration should take only a few seconds, ruling out iterative optimizers (e.g. gradient descent, etc.). In the remainder of this section, we omit the time subscripts from the distributions to lighten the notation; for example $p_0(\mathbf{x}_0) = p(\mathbf{x}_0)$.

\subsection{Simulation-free calibration strategy}

Our key observation is that replacing $\mathbf{u}_{\boldsymbol{\theta}}$ with Eq.~\ref{eq:additive} in Eq.~\ref{eq:loss_unif} yields a linear least-squares problem w.r.t. the guidance weights $\mathbf{w}_{t_k}=\left[w_{u,t_k},w_{f,t_k}\right]^\top$:
\begin{align}
    \mathcal{L}_{\text{free},t_k}(\mathbf{w}_{t_k})&=\mathbb{E}_{p(\mathbf{x}_{t_k}|\mathbf{x}_0)p(\mathbf{y}|\mathbf{x}_0)p(\mathbf{x}_0)}\left[ \left\Vert \mathbf{e}(\mathbf{x}_{t_k},\mathbf{w}_{t_k},\mathbf{x}_0,\mathbf{y},{t_k})\right\Vert^2 \right]
    \label{eq:calib_sim_free}
\end{align}
%\begin{align}
%    \mathcal{L}_{\text{free},t_k}(\mathbf{w}_{t_k})&= \mathbb{E}_{{p}(\mathbf{x}_{0},\mathbf{x}_{t_1},...,\mathbf{x}_{t_K},\mathbf{y})}\left[ \left\Vert \mathbf{e}(\mathbf{x}_{t_k},\mathbf{w}_{t_k},\mathbf{x}_0,\mathbf{y},{t_k})\right\Vert^2 \right]\label{eq:calib_sim_free_joint}\\
%    &=\mathbb{E}_{{p}(\mathbf{x}_{0},\mathbf{x}_{t_k},\mathbf{y})}\left[ \left\Vert \mathbf{e}(\mathbf{x}_{t_k},\mathbf{w}_{t_k},\mathbf{x}_0,\mathbf{y},{t_k})\right\Vert^2 \right]\\
%    &=\mathbb{E}_{p(\mathbf{x}_{t_k}|\mathbf{x}_0)p(\mathbf{y}|\mathbf{x}_0)p(\mathbf{x}_0)}\left[ \left\Vert \mathbf{e}(\mathbf{x}_{t_k},\mathbf{w}_{t_k},\mathbf{x}_0,\mathbf{y},{t_k})\right\Vert^2 \right]
%    \label{eq:calib_sim_free}
%\end{align}
where
\begin{equation}
  \mathbf{e}\left(\mathbf{x_t},\mathbf{w}_t,\mathbf{x}_0,\mathbf{y},t\right)    = 
    {w}_{u,t}\,\mathbf{u}_{\boldsymbol{\theta}}(\mathbf{x}_t,t) + {w}_{f,t}\,\mathbf{f}(\mathbf{x}_t,\mathbf{y},t) - \boldsymbol{\psi}_{t}\left(\mathbf{x}_0,\frac{\mathbf{x}_t - \alpha_t \mathbf{x}_0}{\sigma_t}\right) 
    \label{eq:def_regerr}
\end{equation}
is the regression error, $p(\mathbf{x}_{t_k}|\mathbf{x}_0)$ corresponds to the forward process Eq.~\ref{eq:path} and $p(\mathbf{y}|\mathbf{x}_0)$ is the measurement likelihood Eq.~\ref{eq:obs}. %\JFG{[Est-ce-que~(12) tel que écrit là, notamment avec ce second argument de $\boldsymbol{\psi}_{t}$, reste valide en Flow-Matching~?]}

This observation enables a simulation-free calibration strategy that does not require sampling model trajectories, in which the time-dependent guidance weights are optimized independently at each of the $K$ time steps. Since only two scalar parameters, $w_{u,t_k}$ and $w_{f,t_k}$, need to be estimated at each time step, the expectation over $p(\mathbf{x}_0)$ can be approximated in practice using a single small minibatch, making the overall calibration procedure highly efficient.
    
However, there is a caveat to the simulation-free calibration strategy described above. When optimizing Eq.~\ref{eq:calib_sim_free} over a sufficiently rich function class, such as a neural network, the minimizer corresponds to the true posterior score in the diffusion case~\citep{batzolis2021conditional}. Similar results can be obtained for the flow-matching case. In contrast, our calibration is restricted to only two parameters, and neither the measurement-consistency term nor the unconditional network is necessarily optimal. Therefore, optimizing the guidance weights independently at each time step in Eq.~\ref{eq:calib_sim_free} does not guarantee that the resulting sampling trajectories will be close to the true posterior trajectories. In the following section, we therefore turn to simulation-based calibration.

\subsection{Simulation-based calibration strategy}
We argue that, since neither the measurement-consistency term nor the unconditional network is necessarily optimal, an efficient calibration strategy should be simulation-based. Indeed, during sampling, at a given timestep $t_k$, a sample $\mathbf{x}_{t_{k}}$ is necessarily drawn from an approximate posterior $\widetilde{p}(\mathbf{x}_{t_{k}}|\mathbf{y})$, rather than from the true posterior $p(\mathbf{x}_{t_{k}}|\mathbf{y})$. This discrepancy should therefore be taken into account when calibrating the guidance weights.
 
To do so, we propose to replace the forward-process density $p(\mathbf{x}_{t_k}|\mathbf{x}_0)$, in Eq.~\ref{eq:calib_sim_free}, with the corresponding approximate posterior $\widetilde{p}(\mathbf{x}_{t_k}|\mathbf{y})$. Importantly, $\widetilde{p}(\mathbf{x}_{t_k}|\mathbf{y})$ depends on the guidance weights $\left\{\mathbf{w}_{t_i}\right\}_{i=k+1\ldots K}$.

We therefore obtain the following simulation-based objective:
\begin{equation}
    \mathcal{L}_{\text{sim},t_k}(\mathbf{w}_{t_k},\mathbf{w}_{t_{k+1}},...,\mathbf{w}_{t_{K}})= \mathbb{E}_{\tilde{p}(\mathbf{x}_{t_k}|\mathbf{y})p(\mathbf{y}|\mathbf{x}_0)p(\mathbf{x}_0)}\left[ \left\Vert \mathbf{e}(\mathbf{x}_{t_k},\mathbf{w}_{t_k},\mathbf{x}_{0},\mathbf{y},{t_k})\right\Vert^2 \right], 
    \label{eq:calib_sim_based}
\end{equation} which is a linear least-squares objective in $\mathbf{w}_{t_k}$, but depends non-linearly on the guidance weights on the guidance weights $\left\{\mathbf{w}_{t_i}\right\}_{i=k+1\ldots K}$. To obtain an efficient calibration algorithm, we adopt a greedy strategy in which only the current guidance weights $\mathbf{w}_{t_k}$ are optimized, while the weights calibrated at previous time steps are kept fixed:
\begin{equation}
    \mathbf{w}^*_{t_k} = \arg \min_{\mathbf{w}_{t_k}} \mathcal{L}_{\text{sim},t_k}(\mathbf{w}_{t_k},\mathbf{w}_{t_{k+1}}^*,...,\mathbf{w}_{t_{K}}^*)\label{eq:simulation_based_greed}.
\end{equation}
Since Eq.~\ref{eq:calib_sim_based} is a linear least-squares problem in $\mathbf{w}_{t_k}$, optimization of Eq.~\ref{eq:simulation_based_greed} reduces to a $2\times 2$ linear system.
The proposed SIMulation-based CAlibration strategy (SIMCA) is detailed in Alg.~\ref{algo:calib}. Importantly, the resulting calibration also depends on the choice of the sampler $\Phi$ (see Eq.~\ref{eq:sampler}).

\begin{algorithm}[t]
\caption{Calibration of the guidance weights $\left\{\mathbf{w}_{t_k}\right\}_{k=1\ldots K}$. Black lines are common to both variants; the simulation-based variant adds the \op{blue} lines and skips the \delop{orange} one.}
\label{algo:calib}
\begin{algorithmic}[1]
%\Require pretrained unconditional network $\mathbf{u}_{\boldsymbol{\theta}}$, measurement-consistency term $\mathbf{f}$, minibatch of data and measurements $\{(\mathbf{x}_{0,i},\mathbf{y}_i)\}_{i=1}^N$, timesteps $\left\{t_k\right\}_{k=1\ldots K}$, scheduling $\left\{(\alpha_{t_k},\sigma_{t_k})\right\}_{k=1\ldots K}$, \op{sampler step $\Phi$}
\Require $\mathbf{u}_{\boldsymbol{\theta}}$, $\mathbf{f}$, minibatch $\{(\mathbf{x}_{0,i},\mathbf{y}_i)\}_{i=1}^N$, schedule $\left\{(t_k,\alpha_{t_k},\sigma_{t_k})\right\}_{k=1\ldots K}$, \op{sampler $\Phi$}
%\Ensure weights $\{(a_{t}^*, b_{t}^*)\}_{t}$
\State \op{$\mathbf{x}^{(i)}_{t_K} \sim \mathcal{N}(\mathbf{0},\mathbf{I})$ for $i=1,...,N$} \Comment \op{Initialize sampling trajectories} %\Comment{\op{same initialization as inference}}
\For{$k = K,...,1$} \Comment{for each timestep}%\Comment{parallel over $k$ in the simulation-free variant, sequential for the simulation-based variant}
  \State $t \leftarrow t_k$; \quad $\mathtt{H}\leftarrow 0$; \quad $\mathbf{b} \leftarrow 0$
  \For{$i = 1,\dots,N$}\Comment{for each sample in the minibatch}
    \State \delop{$\boldsymbol{\epsilon}_i \sim \mathcal{N}(\mathbf{0},\mathbf{I})$ \quad $\mathbf{x}^{(i)}_t \leftarrow \alpha_t\mathbf{x}_{0,i} + \sigma_t\boldsymbol{\epsilon}_i$} %\Comment{forward path}
    \State \op{$\boldsymbol{\epsilon}_i \leftarrow (\mathbf{x}^{(i)}_{t} - \alpha_t\mathbf{x}_{0,i})/\sigma_t$} %\Comment{\op{$\mathbf{x}^{(i)}_t$ carried from the previous step}}
    \State $\boldsymbol{\psi}_i \leftarrow \boldsymbol{\psi}_t(\mathbf{x}_{0,i}, \boldsymbol{\epsilon}_i)$ %\Comment{conditional target, Eq.~\eqref{eq:cond_targets}}
    \quad $\mathbf{u}_i \leftarrow \mathbf{u}_{\boldsymbol{\theta}}(\mathbf{x}^{(i)}_t, t)$ \quad $\mathbf{f}_i \leftarrow \mathbf{f}(\mathbf{x}^{(i)}_t, \mathbf{y}_i, t)$
    \State $\mathtt{H} \leftarrow \mathtt{H} +
\begin{bmatrix}
\mathbf{u}_i^\top \mathbf{u}_i & \mathbf{u}_i^\top \mathbf{f}_i \\
\mathbf{f}_i^\top \mathbf{u}_i & \mathbf{f}_i^\top \mathbf{f}_i
\end{bmatrix}
$ \quad $
\mathbf{b} \leftarrow \mathbf{b} +
\begin{bmatrix}
\mathbf{u}_i^\top \boldsymbol{\psi}_i \\
\mathbf{f}_i^\top \boldsymbol{\psi}_i
\end{bmatrix}$
    %\State accumulate $G_{uu} \mathrel{+}= \|\mathbf{u}_i\|^2$, $G_{uf} \mathrel{+}= \mathbf{u}_i^\top\mathbf{f}_i$, $G_{ff} \mathrel{+}= \|\mathbf{f}_i\|^2$, $c_u \mathrel{+}= \mathbf{u}_i^\top\boldsymbol{\psi}_i$, $c_f \mathrel{+}= \mathbf{f}_i^\top\boldsymbol{\psi}_i$
  \EndFor
  %\State $(a^*_{t}, b^*_{t}) \leftarrow$ Eq.~\eqref{eq:closed_form}
    \State $\mathbf{w}^*_{t} \leftarrow$ solve $\mathtt{H}\mathbf{w}= \mathbf{b}$ \Comment 2x2 linear system

  \State \op{$\mathbf{x}^{(i)}_{t_{k+1}} \leftarrow \Phi\big(\mathbf{x}^{(i)}_{t},\, w^*_{u,t}\mathbf{u}_i + w^*_{f,t}\mathbf{f}_i,\, t\big)$ for $i=1,...,N$} \Comment{\op{propagation of sampled trajectories}}
\EndFor
\State \Return $\left\{\mathbf{w}_{t_k}^*\right\}_{k=1\ldots K}$%$\{(a^*_{t_k}, b^*_{t_k})\}_{k}$
\end{algorithmic}
\end{algorithm}

\subsection{Practical considerations}

\paragraph{Cost.} SIMCA requires only a single minibatch of sampling trajectories, making it practical for tuning time-dependent guidance weights for a given measurement setup. Once calibrated, the guidance weights are fixed and introduce no inference-time overhead.

\paragraph{Recalibration.} The guidance weights depend on the measurement operator $\mathcal{A}$, the noise level $\sigma_m$, the measurement-consistency term $\mathbf{f}$, the pretrained unconditional network $\mathbf{u}_{\boldsymbol{\theta}}$, the sampler $\Phi$, and the number of timesteps. They must therefore be recomputed whenever any of these components changes. Since each recalibration requires only one offline pass, practitioners can efficiently adapt the guidance weights to different measurement setups.

\paragraph{Compatibility.} Our calibration procedure applies to any method whose guidance takes the additive form in Eq.~\eqref{eq:additive}. It requires no retraining, architectural changes, or modifications to the sampler.

\section{Experiments} 

\subsection{Experimental Setup}
\paragraph{Backbones and protocols.}
To fairly evaluate SIMCA, we follow the evaluation protocols of DAPS~\citep{zhang2025improving} for diffusion and Flower~\citep{pourya2026flower} for flow matching. See Appendix~\ref{app:exp_details} for further details.
%\paragraph{Backbones and protocols.}
%We evaluate calibration on both families, following an established benchmark in each case so that baseline numbers are directly comparable. 
%For diffusion, we use the pixel-space DDPM checkpoints for FFHQ and ImageNet at $256\times256$ released by~\citep{chung2022diffusion}, and adopt the protocol of DAPS~\citep{zhang2025improving}. For flow matching, we adopt the setup of Flower~\citep{pourya2026flower}. % the same U-Net velocity networks, trained with mini-batch OT flow matching on CelebA and AFHQ-Cat, and the same operators, test images, and evaluation protocol. 
%We evaluate both the OT-coupled network and its independently coupled variant, so that calibration is tested against two different pretrained fields. %All networks are frozen: only the weights $(a_t^*, b_t^*)$ are estimated, and only offline.

\paragraph{Baselines.}
For diffusion we compare against DPS~\citep{chung2022diffusion}, DDRM~\citep{kawar2022denoising}, DDNM~\citep{wang2022zero}, DiffPIR~\citep{zhu2023denoising}, %DCDP~\citep{li2024decoupled},
FPS-SMC~\citep{dou2024diffusion}, and DAPS~\citep{zhang2025improving}. For flow matching we compare against PnP-GS~\citep{hurault2022gradient}, DiffPIR~\citep{zhu2023denoising}, OT-ODE~\citep{pokle2024training}, D-Flow~\citep{benhamu2024dflow}, Flow-Priors~\citep{zhang2024flow}, PnP-Flow~\citep{martin2025pnp}, and Flower~\citep{pourya2026flower}. All baseline metrics are those published by the respective benchmarks, obtained under the same operators, noise levels, and test images as ours, with hyperparameters grid-searched to maximize PSNR. %We compare against the single-evaluation variants PnP-Flow$^1$ and Flower$^1$, since SIMCA likewise uses one velocity evaluation per step.

\paragraph{Calibration protocol.}
Guidance weights are calibrated offline on images disjoint from the test images: for flow matching, we use the same calibration split as in~\citep{pourya2026flower}; % the same split the benchmark uses to grid-search its baselines, so our method sees no more data; 
for diffusion, we use images from the training sets of FFHQ~\citep{karras2019style} and ImageNet~\citep{deng2009imagenet}. All main tables report SIMCA, the simulation-free variant being deferred to Section~\ref{sec:ablations}. Since the weights are tied to the sampler and its discretization, each configuration is calibrated separately and its step count reported: $K \in \{1000, 500, 50\}$ for diffusion, with a finer sweep in Appendix~\ref{app:samplers_ablation}, and the renoising sampler of Flower (with the same number of steps) or a plain Euler integrator for flow matching.

\subsection{Main results}
\label{sec:quant}

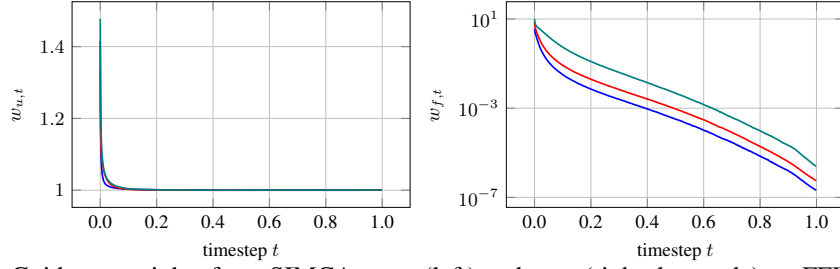
\begin{figure}[t]
    \centering
    \resizebox{0.8\linewidth}{!}{%
    \begin{tikzpicture}
\begin{groupplot}[
    group style={
        group size=2 by 1,
        horizontal sep=1.6cm,
    },
    width=7.2cm,
    height=5cm,
    xlabel={timestep $t$},
    xticklabels={0.1, 0.0, 0.2, 0.4, 0.6, 0.8, 1.0},
    grid=major,
    tick label style={font=\small},
    label style={font=\small},
    cycle list={
        {blue,thick},
        {red,thick},
        {teal,thick},
    },
]

% --- left panel: a_p ---
\nextgroupplot[ylabel={$w_{u,t}$}]
\addplot table [x=t, y=a_p, col sep=comma] {a_star_weights_1000_100_box_inpainting.csv};
\addplot table [x=t, y=a_p, col sep=comma] {a_star_weights_1000_100_random_inpainting.csv};
\addplot table [x=t, y=a_p, col sep=comma] {a_star_weights_1000_100_super_resolution.csv};

% --- right panel: a_l (log scale) ---
\nextgroupplot[ylabel={$w_{f,t}$}, ymode=log]
\addplot table [x=t, y=a_l, col sep=comma] {a_star_weights_1000_100_box_inpainting.csv};
\addplot table [x=t, y=a_l, col sep=comma] {a_star_weights_1000_100_random_inpainting.csv};
\addplot table [x=t, y=a_l, col sep=comma] {a_star_weights_1000_100_super_resolution.csv};

\end{groupplot}
\end{tikzpicture}}
    \vspace{-1.0\baselineskip}
\caption{Guidance weights from SIMCA: $w_{u,t}$ (left) and $w_{f,t}$ (right, log scale) on FFHQ at $K=1000$, for \textcolor{blue}{box inpainting}, \textcolor{red}{random inpainting} and \textcolor{teal}{super-resolution}.}
    \label{fig:weights}
\end{figure}

\paragraph{Calibrated guidance weights.}
Fig.~\ref{fig:weights} shows the guidance weights obtained on FFHQ for three measurement operators. The schedules are therefore neither constant nor reproducible by a single hand-designed curve shared across operators, which is what the remainder of this section quantifies.

\paragraph{Diffusion backbones.}
Table~\ref{tab:quant_results} reports results on FFHQ and ImageNet. SIMCA either matches or outperforms the state-of-the-art methods on almost every metric, task, and dataset: on random inpainting, LPIPS drops from $0.203$ to $0.068$ and FID from $69.20$ to $18.31$ on FFHQ, and from $0.297$ to $0.109$ and $87.53$ to $32.08$ on ImageNet.

\begin{figure}[t]
    \centering

    \begin{subfigure}[b]{0.48\linewidth}
        \centering
        \includegraphics[width=0.9\linewidth]{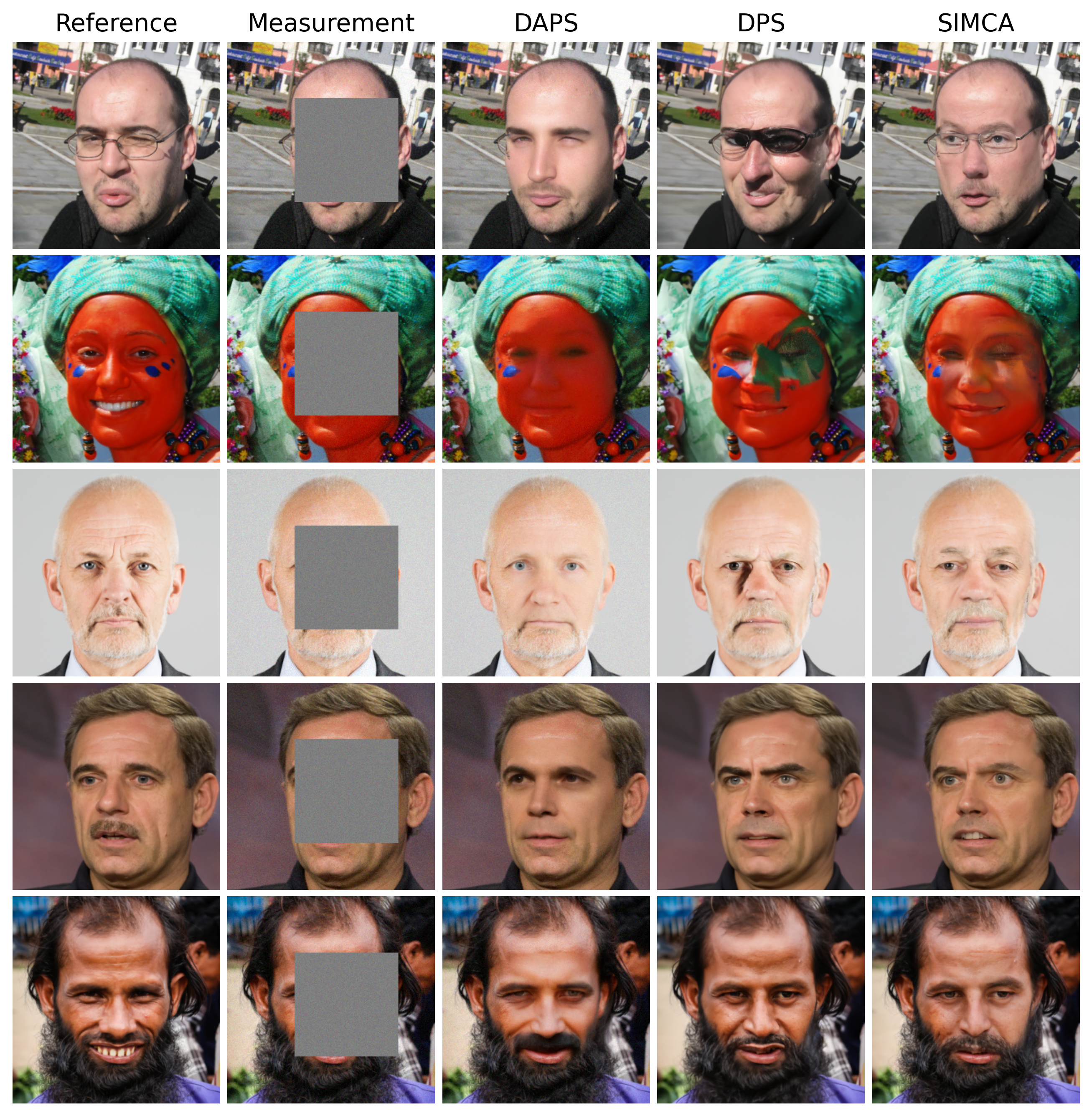}
        \caption{FFHQ, diffusion backbone.}
        \label{fig:figure1}
    \end{subfigure}
    \hfill
    \begin{subfigure}[b]{0.48\linewidth}
        \centering
        \includegraphics[width=0.9\linewidth]{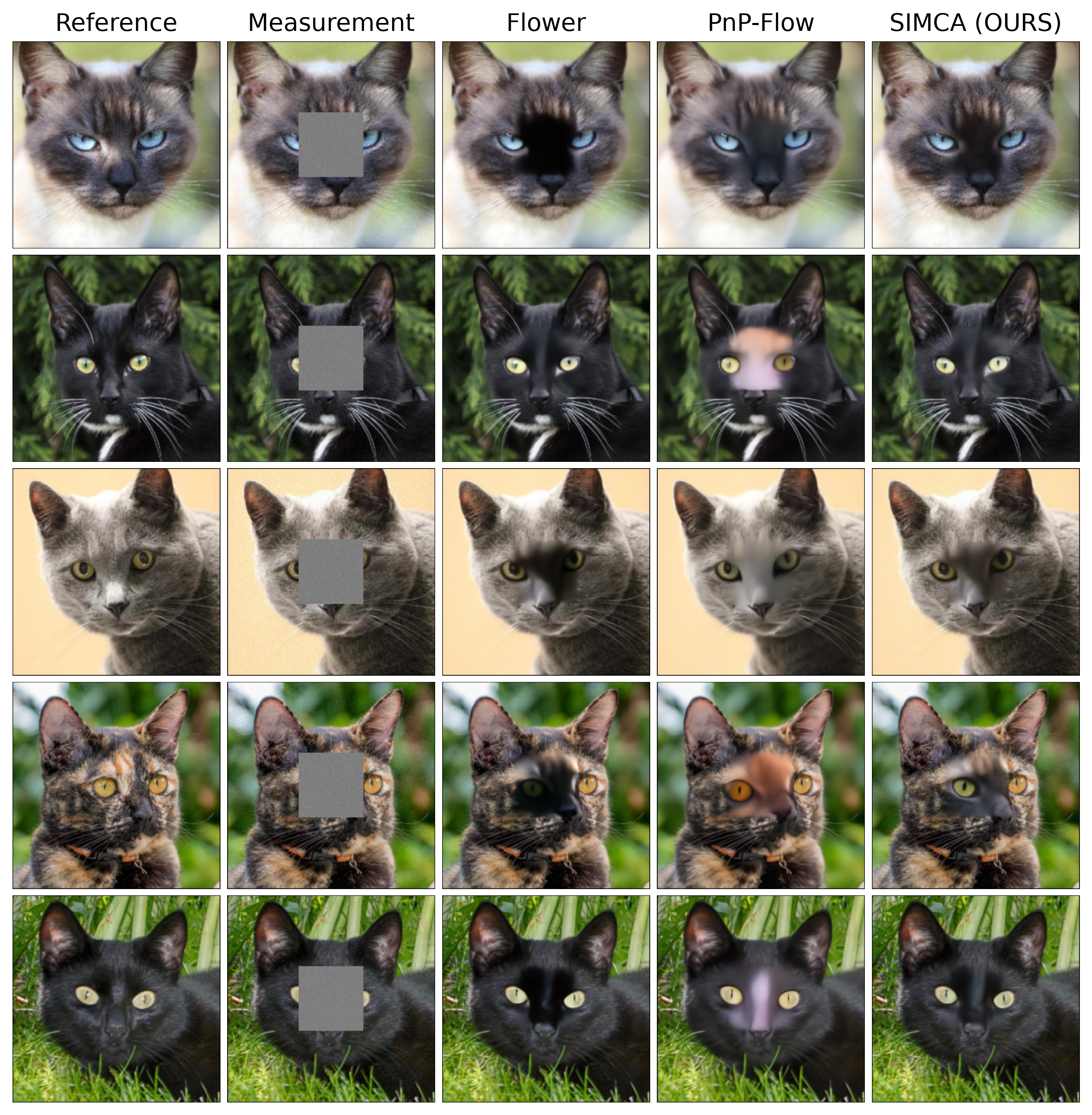}
        \caption{AFHQ-Cat, flow-matching backbone.}
        \label{fig:figure2}
    \end{subfigure}

    \caption{Qualitative comparison on the box inpainting task. SIMCA produces completions coherent with the visible context, whereas the baselines oversmooth the fill, 
    %(DAPS, (a)), 
    hallucinate details, 
    %(DPS, sunglasses in (a), first row),
    leak colours, 
    %(DPS, (a), second row), 
    or produce dark or discoloured fills.% (Flower and PnP-Flow, (b)).
    }
    \label{fig:qual_box}
\end{figure}

The most informative comparison is against DAPS~\citep{zhang2025improving}, the strongest and most elaborate baseline. SIMCA surpasses it on random inpainting by a wide margin on both datasets ($31\%$ lower LPIPS and $43\%$ lower FID on FFHQ), on all four metrics on FFHQ super-resolution, and on PSNR, SSIM, and LPIPS on FFHQ motion deblurring. On box inpainting the results are mixed: SSIM and LPIPS improve over every baseline, while PSNR remains below DDNM and FPS-SMC. This is the signature of a sampler that explores the posterior rather than collapsing onto its mean, and box inpainting is the task with the broadest posterior. Figure~\ref{fig:qual_box} shows this directly: SIMCA completes the structure visible outside the mask, where DAPS oversmooths and DPS hallucinates.

\setlength{\tabcolsep}{2pt} %<<<==
\begin{table*}[t]
  \caption{Reconstruction quality on FFHQ and ImageNet. Within each task, best values are in \textbf{Bold} and second best are \underline{underlined}. SIMCA-$K$ denotes calibration and sampling with $K$ steps.}
  \label{tab:quant_results}
  \centering
  \footnotesize
  \setlength{\tabcolsep}{3pt}
  \renewcommand{\arraystretch}{0.92}
  \resizebox{0.7\textwidth}{!}{%
  \begin{tabular}{c|c|cccc|cccc}
    \toprule
    \multirow{2}{*}{Task} & \multirow{2}{*}{Method} & \multicolumn{4}{c|}{FFHQ} & \multicolumn{4}{c}{ImageNet} \\
    & & PSNR$\uparrow$ & SSIM$\uparrow$ & LPIPS$\downarrow$ & FID$\downarrow$ & PSNR$\uparrow$ & SSIM$\uparrow$ & LPIPS$\downarrow$ & FID$\downarrow$ \\
    \midrule
    \multirow{10}{*}{SR$\times 4$} & DAPS & 29.07 & 0.818 & 0.177 & 51.44 & 25.89 & 0.694 & 0.276 & \textbf{83.57} \\
    & DPS & 25.86 & 0.753 & 0.269 & 81.07 & 21.13 & 0.489 & 0.361 & 106.32\\
    & DDRM & 26.58 & 0.782 & 0.282 & 79.25 & 22.62 & 0.521 & 0.324 & 103.85 \\
    & DDNM & 28.03 & 0.795 & 0.197 & 64.62 & 23.96 & 0.604 & 0.475 & 98.62\\
    %& DCDP & 28.66 & 0.807 & 0.178 & 53.81 & - & - & - & -\\
    & FPS-SMC & 28.42 & 0.813 & 0.204 & 49.25 & 24.82 & 0.703 & 0.313 & 97.51\\ 
    & DiffPIR & 26.64 & - & 0.260 & 65.77 & 23.18 & - & 0.371 & 106.32\\ 
    % \cdashline{2-10}
    %& DPS & 28.12 & 0.795 & 0.169 & 53.51 & 25.129 & 0.659 & 0.265 & 97.74\\
    & SIMCA-1k & 29.12 & 0.821 & 0.145 & \underline{46.39} & 26.14 & 0.712 & \underline{0.241} & 97.48 \\
    & SIMCA-500 & \underline{29.26} & \underline{0.826} & \textbf{0.143} & \textbf{43.14} & \underline{26.17} & \underline{0.715} & \textbf{0.239} & \underline{93.57} \\
    & SIMCA-50 & \textbf{30.00} & \textbf{0.853} & \underline{0.144} & 49.02 & \textbf{26.45} & \textbf{0.726} & 0.244 & 107.25 \\
    \midrule
    \multirow{9}{*}{\shortstack{Box\\Inpainting}} & DAPS & 24.07 & 0.814 & 0.133 & \underline{43.10} & 21.43 & 0.725 & 0.214 & \underline{109.85}\\
    & DPS & 22.51 & 0.792 & 0.209 & 61.27 & 18.94 & 0.722 & 0.257 & 126.52\\
    & DDRM & 22.26 & 0.801 & 0.207 & 78.62 & 18.63 & 0.733 & 0.254 & 116.37\\
    & DDNM & \underline{24.47} & 0.837 & 0.235 & 46.59 & \underline{21.64} & 0.748 & 0.319 & \textbf{103.97}\\
    %& DCDP & 23.89 & 0.760 & 0.163 & 45.23 & - & - & - & -\\
    & FPS-SMC & \textbf{24.86} & 0.823 & 0.146 & 48.34 & \textbf{22.16} & 0.726 & 0.208 & 111.58\\
    % \cdashline{2-10}
    %& DPS & 23.75 & 0.845 & 0.122 & 47.06 & 19.39 & 0.748 & 0.217 & 115.70\\
    & SIMCA-1k & 22.90 & 0.846 & \underline{0.106} & 43.68 & 19.56 & \underline{0.786} & \textbf{0.185} & 135.29 \\
    & SIMCA-500 & 23.28 & \underline{0.850} & \textbf{0.105} & \textbf{39.77} & 19.63 & \textbf{0.789} & \underline{0.186} & 140.73 \\
    & SIMCA-50 & 23.42 & \textbf{0.855} & 0.110 & 47.12 & 19.86 & 0.778 & 0.210 & 146.87 \\
    \midrule
    \multirow{8}{*}{\shortstack{Random\\Inpainting}} & DAPS & 31.12 & 0.844 & 0.098 & 32.17 & 28.44 & 0.775 & 0.135 & 54.25\\
    & DPS & 25.46 & 0.823 & 0.203 & 69.20 & 23.52 & 0.745 & 0.297 & 87.53\\
    & DDNM & 29.91 & 0.817 & 0.121 & 44.37 & \textbf{31.16} & 0.841 & 0.191 & 63.84\\
    %& DCDP & 30.69 & 0.842 & 0.142 & 52.51 & - & - & - & -\\
    & FPS-SMC & 28.21 & 0.823 & 0.261 & 61.23 & 24.52 & 0.701 & 0.316 & 79.12\\
    % \cdashline{2-10}
    %& DPS & 31.54 & 0.888 & 0.120 & 42.06 & 28.18 & 0.788 & 0.193 & 67.41\\
    & SIMCA-1k & 32.91 & 0.916 & \underline{0.069} & \underline{19.16} & 29.22 & 0.851 & \textbf{0.109} & \textbf{32.08} \\
    & SIMCA-500 & \underline{33.07} & \underline{0.920} & \textbf{0.068} & \textbf{18.31} & 29.30 & \textbf{0.853} & \underline{0.110} & \underline{33.69} \\
    & SIMCA-50 & \textbf{33.76} & \textbf{0.930} & 0.072 & 24.88 & \underline{29.46} & \underline{0.852} & 0.132 & 46.08 \\
    \midrule
    \multirow{10}{*}{\shortstack{Gaussian\\Deblurring}} & DAPS & \textbf{29.19} & \textbf{0.817} & 0.165 & \textbf{53.33} & \underline{26.15} & \underline{0.684} & \textbf{0.253} & \textbf{75.68}\\
    & DPS & 25.87 & 0.764 & 0.219 & 79.75 & 20.31 & 0.598 & 0.397 & 116.42 \\
    & DDRM & 24.93 & 0.732 & 0.239 & 92.43 & 21.26 & 0.564 & 0.443 & 146.89\\
    & DDNM & 28.20 & 0.804 & 0.216 & 57.83 & \textbf{28.06} & \textbf{0.703} & 0.278 & \underline{81.43}\\
    %& DCDP & 27.50 & 0.699 & 0.304 & 86.43 & - & - & - & - \\
    & FPS-SMC & 26.54 & 0.773 & 0.253 & 67.45 & 23.91 & 0.601 & 0.387 & 91.72\\
    & DiffPIR & 27.36 & - & 0.236 & 59.65 & 22.80 & - & 0.355 & 93.36\\
    % \cdashline{2-10}
    %& DPS & 27.29 & 0.767 & \underline{0.165} & \textbf{49.97} & 20.53 & 0.477 & 0.354 & 144.12 \\
    & SIMCA-1k & 28.23 & 0.794 & \underline{0.163} & 57.07 & 25.28 & 0.668 & 0.281 & 121.85 \\
    & SIMCA-500 & 28.29 & 0.798 & \textbf{0.161} & \underline{53.60} & 25.19 & 0.672 & \underline{0.272} & 120.68 \\
    & SIMCA-50 & \underline{28.33} & \underline{0.812} & 0.165 & 58.54 & 25.02 & 0.663 & 0.294 & 141.21 \\
    \midrule
    \multirow{8}{*}{\shortstack{Motion\\Deblurring}} & DAPS & 29.66 & 0.847 & 0.157 & \textbf{39.49} & \textbf{27.86} & 0.766 & \textbf{0.196} & \textbf{61.83} \\
    & DPS & 24.52 & 0.801 & 0.246 & 65.23 & 18.96 & 0.629 & 0.423 & 137.81\\
    %& DCDP & 25.08 & 0.512 & 0.364 & 125.13 & - & - & - & -\\
    & FPS-SMC & 27.39 & 0.826 & 0.227 & 48.32 & 24.52 & 0.647 & 0.326 & 87.43\\
    & DiffPIR & 26.57 & - & 0.255 & 65.78 & 24.01 & - & 0.366 & 94.63\\
    % \cdashline{2-10}
    %& DPS & 29.35 & 0.827 & 0.131 & 43.90 & 24.72 &0.652& 0.261 & 95.33\\
    & SIMCA-1k & \underline{30.55} & 0.850 & \underline{0.122} & 43.22 & \underline{27.67} & \underline{0.771} & \underline{0.210} & \underline{83.00} \\
    & SIMCA-500 & \textbf{30.64} & \underline{0.855} & \textbf{0.121} & \underline{40.45} & \underline{27.67} & \textbf{0.774} & 0.211 & 86.42 \\
    & SIMCA-50 & 30.29 & \textbf{0.863} & 0.131 & 43.60 & 26.77 & 0.753 & 0.236 & 112.84 \\
    \bottomrule
  \end{tabular}
  }
\end{table*}

\paragraph{Effect of the step count.}
The three SIMCA rows in each block of Table~\ref{tab:quant_results} differ only in the number of sampler iterations. Surprisingly, halving the budget from $1000$ to $500$ steps improves almost every metric on every task. Reducing further to $50$ improves PSNR and SSIM again, reaching the best values in the table on four of the five FFHQ tasks, while LPIPS and FID degrade moderately: coarser discretization accumulates less stochasticity, moving the sampler toward the posterior mean, which distortion metrics reward and perceptual metrics penalize. A $20\times$ reduction in inference cost is therefore available at a bounded perceptual cost, and a $2\times$ reduction at none.%, provided the weights are recalibrated at the target discretization.

\paragraph{Flow matching backbones.}
Tables~\ref{tab:celeba} and~\ref{tab:afhq_cat} report results on CelebA and AFHQ-Cat, where the pretrained field is a velocity rather than a score. SIMCA is applied unchanged, which is itself a strong result as it bridges the gap between training-free diffusion and flow-matching methods.%: the regression error of Eq.~\ref{eq:def_regerr} refers only to a pretrained field and a measurement-consistency term, so neither the derivation nor the implementation distinguishes the two families.

On CelebA, SIMCA attains the best value on fourteen of the fifteen columns, improving on the strongest baseline, Flower~\citep{pourya2026flower}, on every task, with PSNR gains of $1.7$\,dB on random inpainting and $1.2$\,dB on box inpainting. On AFHQ-Cat the gains are more nuanced: SIMCA leads on both inpainting tasks across all three metrics (Figure~\ref{fig:qual_box}) and on super-resolution PSNR and SSIM, while PnP-GS, Flower and OT-ODE perform best on denoising and deblurring. We refer the reader to Appendix~\ref{app:robust} for a deeper analysis.%SIMCA perform the best on both datasets precisely on the inpainting problems, where the operator induces significant signal loss and the balance between prior and measurement information matters most, and not where the operator is well conditioned and the baselines already agree closely.

\begin{table*}[t]
\centering
\caption{Results on 100 test images of the dataset CelebA.}
\label{tab:celeba}
\footnotesize
\resizebox{0.9\textwidth}{!}{%
\begin{tabular}{l ccc ccc ccc ccc ccc}
\toprule
& \multicolumn{3}{c}{Denoising} & \multicolumn{3}{c}{Deblurring}
& \multicolumn{3}{c}{Super-resolution} & \multicolumn{3}{c}{Random inpainting}
& \multicolumn{3}{c}{Box inpainting} \\
\cmidrule(lr){2-4} \cmidrule(lr){5-7} \cmidrule(lr){8-10} \cmidrule(lr){11-13} \cmidrule(lr){14-16}
Method & PSNR & SSIM & LPIPS & PSNR & SSIM & LPIPS & PSNR & SSIM & LPIPS
& PSNR & SSIM & LPIPS & PSNR & SSIM & LPIPS \\
\midrule
%Degraded
% & 20.00 & 0.348 & 0.372 & 27.83 & 0.740 & 0.126 & 10.26 & 0.183 & 0.827 & 11.95 & 0.196 & 1.041 & 22.27 & 0.742 & 0.214 \\
PnP-GS
& \textbf{32.64} & 0.910 & 0.035 & 34.03 & 0.924 & 0.041 & 31.31 & 0.892 & 0.064 & 29.22 & 0.875 & 0.070 & -- & -- & -- \\
DiffPIR
& 31.20 & 0.885 & 0.060 & 32.77 & 0.912 & 0.060 & 31.52 & 0.895 & 0.033 & 31.74 & 0.917 & 0.025 & -- & -- & -- \\
OT-ODE
& 30.54 & 0.859 & 0.032 & 33.01 & 0.921 & 0.029 & 31.46 & 0.907 & 0.025 & 28.68 & 0.871 & 0.051 & 29.40 & 0.920 & 0.038 \\
D-Flow
& 26.04 & 0.607 & 0.092 & 31.25 & 0.854 & 0.038 & 30.47 & 0.843 & 0.026 & 33.67 & 0.943 & 0.015 & 30.70 & 0.899 & 0.026 \\
Flow-Priors
& 29.34 & 0.768 & 0.134 & 31.54 & 0.858 & 0.056 & 28.35 & 0.713 & 0.102 & 32.88 & 0.871 & 0.019 & 30.07 & 0.858 & 0.048 \\
PnP-Flow
& 31.80 & 0.905 & 0.044 & 34.48 & 0.936 & 0.040 & 31.09 & 0.902 & 0.045 & 33.05 & 0.944 & 0.018 & 30.47 & 0.933 & 0.037 \\
Flower-OT
& 32.28 & \underline{0.914} & 0.034 & \underline{34.98} & \textbf{0.947} & 0.026 & 32.36 & \underline{0.923} & 0.034 & 33.08 & 0.944 & 0.018 & 31.19 & 0.945 & 0.022 \\
\midrule
SIMCA (ren.) &
\textbf{32.64} & \textbf{0.919} & \textbf{0.020} & \textbf{35.17} & \textbf{0.947} & \underline{0.023} & \textbf{33.02} & \textbf{0.929} & \underline{0.021} & \textbf{34.80} & \textbf{0.962} & \textbf{0.012} & \textbf{32.40} & \textbf{0.955} & \textbf{0.018} \\
SIMCA + OT (ren.) &
\underline{32.42} & \underline{0.914} & \underline{0.021} &
34.87 & \underline{0.943} & \textbf{0.022} &
\underline{32.89} & \textbf{0.929} & \textbf{0.019} &
\underline{34.40} & \underline{0.960} & \underline{0.013} &
\underline{31.77} & \underline{0.95}1 & \underline{0.020} \\
SIMCA (eul.) & 
31.83 & 0.897 & 0.026 &
33.30 & 0.927 & 0.031 &
31.45 & 0.910 & 0.025 &
31.10 & 0.928 & 0.022 &
29.45 & 0.935 & 0.026 \\
SIMCA + OT (eul.) & 
31.65 & 0.895 & 0.028 &
33.09 & 0.924 & 0.033 &
31.26 & 0.908 & 0.026 &
30.77 & 0.924 & 0.025 &
28.96 & 0.933 & 0.028 \\
\bottomrule
\end{tabular}
}
\end{table*}

\begin{table*}[t]
\centering
\caption{Results on 100 test images of the dataset AFHQ-Cat.}
\label{tab:afhq_cat}
\footnotesize
\resizebox{0.9\textwidth}{!}{%
\begin{tabular}{l ccc ccc ccc ccc ccc}
\toprule
& \multicolumn{3}{c}{Denoising} & \multicolumn{3}{c}{Deblurring}
& \multicolumn{3}{c}{Super-resolution} & \multicolumn{3}{c}{Random inpainting}
& \multicolumn{3}{c}{Box inpainting} \\
\cmidrule(lr){2-4} \cmidrule(lr){5-7} \cmidrule(lr){8-10} \cmidrule(lr){11-13} \cmidrule(lr){14-16}
Method & PSNR & SSIM & LPIPS & PSNR & SSIM & LPIPS & PSNR & SSIM & LPIPS
& PSNR & SSIM & LPIPS & PSNR & SSIM & LPIPS \\
\midrule
% Degraded
% & 20.00 & 0.314 & 0.509 & 23.94 & 0.517 & 0.444 & 11.70 & 0.208 & 0.873 & 13.36 & 0.223 & 1.081 & 21.80 & 0.740 & 0.198 \\
PnP-GS
& \textbf{32.58} & \textbf{0.894} & 0.072 & 27.91 & 0.753 & 0.349 & 24.15 & 0.632 & 0.362 & 29.42 & 0.836 & 0.126 & -- & -- & -- \\
DiffPIR
& 30.58 & 0.835 & 0.189 & 27.56 & 0.728 & 0.342 & 23.65 & 0.624 & 0.402 & 31.70 & 0.881 & 0.062 & -- & -- & -- \\
OT-ODE
& 30.03 & 0.815 & 0.076 & 27.06 & 0.713 & \textbf{0.123} & 25.91 & 0.716 & \textbf{0.108} & 29.40 & 0.839 & 0.090 & 24.62 & 0.875 & 0.085 \\
D-Flow
& 26.13 & 0.574 & 0.175 & 27.82 & 0.721 & \underline{0.164} & 24.64 & 0.601 & 0.190 & 32.20 & 0.894 & 0.040 & 26.26 & 0.842 & 0.077 \\
Flow-Priors
& 29.41 & 0.763 & 0.153 & 26.47 & 0.700 & 0.181 & 23.51 & 0.570 & 0.272 & 32.37 & 0.906 & 0.047 & 26.20 & 0.818 & 0.118 \\
PnP-Flow
& 31.18 & 0.863 & 0.135 & 27.87 & 0.760 & 0.304 & \underline{26.94} & \underline{0.763} & 0.171 & 33.00 & 0.918 & 0.037 & 26.00 & 0.897 & 0.103 \\
Flower-OT
& \underline{31.69} & 0.879 & 0.102 & \textbf{28.64} & \textbf{0.775} & 0.255 & 26.23 & 0.741 & 0.272 & 32.97 & 0.918 & 0.040 & 26.19 & 0.915 & 0.063 \\
\midrule
SIMCA (ren.)
& 31.58 & \underline{0.880} & \textbf{0.054} & 27.69 & 0.757 & 0.245 & \textbf{27.18} & \textbf{0.776} & 0.169 & \textbf{33.53} & \textbf{0.926} & \textbf{0.031} & \textbf{27.46} & \textbf{0.921} & \textbf{0.049} \\
SIMCA + OT (ren.)
& 31.53 & 0.876 & 0.061 & \underline{28.12} & \underline{0.767} & 0.269 & 24.29 & 0.754 & 0.205 & \underline{33.34} & \underline{0.924} & \underline{0.036} & \underline{26.87} & \underline{0.919} & 0.058 \\
SIMCA (eul.)
& 31.40 & 0.864 & \underline{0.060} & 26.64 & 0.680 & 0.169 & 26.14 & 0.710 & \underline{0.125} & 31.57 & 0.894 & 0.039 & 24.85 & 0.899 & \underline{0.055} \\
SIMCA + OT (eul.)
& 31.44 & 0.864 & 0.060 & 26.47 & 0.668 & 0.174 & 25.98 & 0.705 & 0.129 & 31.55 & 0.893 & 0.042 & 24.47 & 0.739 & 0.058 \\
\bottomrule
\end{tabular}
}
\end{table*}

% \paragraph{Robustness to the backbone and the sampler.}
% Two comparisons isolate what the gains depend on. First, in the flow-matching experiments, the OT-coupled and independently coupled backbones behave almost identically once calibrated, differing by at most $0.2$\,dB and $0.002$ LPIPS on CelebA. %Therefore, SIMCA 
% %calibration
% %does not rely on a particular pretrained field, and in particular not on the OT-coupling the strongest baselines are built on. 
% Second, the renoising sampler outperforms the plain Euler integrator by $3.7$\,dB on CelebA random inpainting and $3.0$\,dB on box inpainting. This matches the deterministic-sampler penalty of Section~\ref{sec:ablations} and is a property of the sampler rather than of calibration: the Euler rows still improve on most flow matching baselines.

% \paragraph{Where SIMCA does not help.} %calibration 
% On ImageNet Gaussian deblurring, SIMCA %calibrated DPS
% trails the strongest baselines on every metric, reaching $25.28$\,dB against $28.06$ for DDNM and $26.15$ for DAPS, and the same pattern appears on AFHQ-Cat deblurring and denoising against PnP-GS and OT-ODE. The improvement over DPS is nonetheless the largest in the table, from $20.31$ to $25.28$\,dB, so the residual gap reflects a Gaussian likelihood that is genuinely limiting on these operators rather than merely mis-weighted. SIMCA
% %Calibration
% recovers the best combination available within the chosen family of measurement terms.
% %the experiments establish that this family is adequate far more often than the recent literature has assumed, not that it always is.

\paragraph{Where SIMCA does not help.}
On ImageNet Gaussian deblurring, SIMCA trails the strongest baselines on every metric, reaching $25.28$\,dB against $28.06$ for DDNM, and the same pattern appears on AFHQ-Cat deblurring and denoising. The improvement over DPS is the largest in the table, from $20.31$ to $25.28$\,dB, so the residual gap reflects a Gaussian likelihood that is genuinely limiting on these operators.% rather than merely mis-weighted.%: SIMCA recovers the best combination available within the chosen family of measurement terms, not beyond it.

\subsection{Ablation studies}
\label{sec:ablations}

\paragraph{Calibration set size.}
Table~\ref{tab:calibration_images} varies the number of calibration images on FFHQ random and box inpainting. A single image already suffices. %: LPIPS is constant to three decimals, and FID varies by less than $0.2$ and $0.5$ respectively across two orders of magnitude. 
This is expected, since only two scalars are fitted per time step while each image contributes a full-dimensional inner product. Calibration is correspondingly cheap: solving for one weight pair takes $0.11$\,s from a single image and $6.76$\,s from one hundred, so a full schedule is obtained in seconds to minutes.

% \begin{table}[ht]
% \centering
% \caption{LPIPS, FID with respect to the number of calibration images.}
% \label{tab:calibration_images}
% \begin{tabular}{ccccc|c}
% \toprule
% \shortstack{Calibration\\images} &
% \shortstack{LPIPS Rand.\\Inpainting}&
% \shortstack{LPIPS Box\\Inpainting} &
% \shortstack{FID Rand.\\Inpainting} &
% \shortstack{FID Box\\Inpainting} &
% \shortstack{Computation time \\ per $(a^*_t, b_t^*)$} \\
% \midrule
% 1   & 0.069 & 0.107 & 19.03 & 44.13 & 0.113 \\
% 2   & 0.069 & 0.106 & 19.14 & 43.84 & 0.180 \\
% 5   & 0.069 & 0.106 & 19.18 & 43.61 & 0.379 \\
% 10  & 0.069 & 0.106 & 19.18 & 43.79 & 0.675 \\
% 20  & 0.069 & 0.106 & 19.20 & 43.62 & 1.352 \\
% 50  & 0.069 & 0.106 & 19.17 & 43.71 & 3.372 \\
% 100 & 0.069 & 0.106 & 19.16 & 43.68 & 6.756 \\
% \bottomrule
% \end{tabular}
% \end{table}

\paragraph{Simulation-free versus simulation-based calibration.}
Table~\ref{tab:ablation_surrogates} compares the two strategies of Section~\ref{sec:method} on FFHQ random inpainting. Both improve substantially on the uncalibrated sampler. SIMCA is nonetheless uniformly better, with the largest margin on the perceptual metrics: for DPS, LPIPS falls from $0.103$ to $0.069$ and FID from $33.71$ to $19.16$, a $43\%$ reduction for the same number of samples, and the same ordering holds for $\Pi$GDM and TMPD.

% \paragraph{Forward versus  simulation-based calibration.} %on-policy
% Table~\ref{tab:ablation_surrogates} compares the two estimators of Section~\ref{sec:estimation} on FFHQ random inpainting. Both improve substantially on the uncalibrated sampler, so most of the gain comes from calibrating at all rather than from the choice of distribution. Simulation-based
% %On-policy
% calibration is nonetheless uniformly better, with the largest margin on the perceptual metrics: for DPS, LPIPS falls from $0.103$ to $0.069$ and FID from $33.71$ to $19.16$, a $43\%$ reduction for no additional samples, and the same ordering holds for $\Pi$GDM and TMPD. This supports the exposure-bias argument: weights fitted on the forward marginals are measurably suboptimal on the states the sampler visits, and the discrepancy appears in distributional rather than distortion metrics.

\paragraph{Choice of the measurement-consistency term.}
Table~\ref{tab:ablation_surrogates} also compares different measurement-consistency terms: the isotropic Gaussian likelihood of DPS which all other experiments are based on (see Eq.~\ref{eq:additive}), the pseudoinverse-based covariance correction of $\Pi$GDM, and the Tweedie moment projection of TMPD. Uncalibrated, they span $25.10$ to $27.89$\,dB and $0.156$ to $0.248$ LPIPS, in the order the literature reports. After simulation-based 
%on-policy
calibration they become nearly indistinguishable: $32.91$, $32.97$, and $32.84$\,dB, with LPIPS of $0.069$, $0.069$, and $0.070$. TMPD is the telling case, leading by $2.4$\,dB uncalibrated and ending within $0.07$\,dB of DPS.

The advantage of a more elaborate surrogate over the isotropic Gaussian one is therefore, in these settings, an advantage in effective scaling rather than in expressiveness, and it almost vanishes after calibration with SIMCA. %The residual error after calibration is common to all three, so further progress requires leaving the additive family rather than refining within it.

% \begin{table*}
%   \caption{Calibration applied to three measurement-consistency terms, on FFHQ random inpainting. Both estimators of Section~\ref{sec:estimation} are reported; "fwd" is simulation-free calibration and "bwd" is SIMCA.}
%   \label{tab:ablation_surrogates}
%   \centering
%   \begin{tabular}{l|cccc}
%     \toprule
%     & PSNR $\uparrow$ & SSIM $\uparrow$ & LPIPS $\downarrow$ & FID $\downarrow$ \\ \bottomrule
%     DPS & $25.46$ & $0.823$ & $0.203$ & $69.20$ \\
%     %DPS + DPS post-process.& $29.59$ & $0.880$ & $0.123$ & $41.22$ & $23.27$ & $0.842$ & $0.120$ & $44.13$ \\
%     %DPS + DAPS post-process.& $31.54$ & $0.888$ & $0.120$ & $42.06$ & $23.75$ & $0.845$ & $0.122$ & $47.06$ \\
%     DPS + calib fwd & $31.69$ & $0.897$ & $0.103$ & $33.71$ \\
%     %DPS + DAPS post-process. + calib bwd & $32.90$ & $0.915$ & $0.069$ & $19.10$ & $22.93$ & $0.846$ & $0.106$ & $45.14$ \\
%     DPS + calib bwd & $32.91$ & $0.916$ & $0.069$ & $19.16$ \\ \midrule
%     $\Pi$GDM & $25.10$ & $0.719$ & $0.248$ & $67.93$ \\
%     $\Pi$GDM + calib fwd & $32.12$ & $0.903$ & $0.096$ & $30.68$ \\
%     $\Pi$GDM + calib bwd & $32.97$ & $0.916$ & $0.069$ & $18.35$ \\ \midrule
%     TMPD & $27.89$ & $0.835$ & $0.156$ & $56.43$ \\
%     TMPD + calib fwd & $32.02$ & $0.879$ & $0.115$ & $30.72$ \\
%     TMPD + calib bwd & $32.84$ & $0.915$ & $0.070$ & $20.21$ \\
%   \end{tabular}
% \end{table*}

\begin{table*}[t]
\centering
\footnotesize

\begin{minipage}[t]{0.55\linewidth}
\centering
\caption{Calibration applied to three measurement-consistency terms, on FFHQ random inpainting.}
\label{tab:ablation_surrogates}
\resizebox{0.6\linewidth}{!}{%
\begin{tabular}{l|cccc}
\toprule
& PSNR & SSIM & LPIPS & FID \\ \midrule
DPS               & 25.46 & 0.823 & 0.203 & 69.20 \\
\quad + sim.-free & 31.69 & 0.897 & 0.103 & 33.71 \\
\quad + SIMCA     & 32.91 & 0.916 & 0.069 & 19.16 \\ \midrule
$\Pi$GDM          & 25.10 & 0.719 & 0.248 & 67.93 \\
\quad + sim.-free & 32.12 & 0.903 & 0.096 & 30.68 \\
\quad + SIMCA     & 32.97 & 0.916 & 0.069 & 18.35 \\ \midrule
TMPD              & 27.89 & 0.835 & 0.156 & 56.43 \\
\quad + sim.-free & 32.02 & 0.879 & 0.115 & 30.72 \\
\quad + SIMCA     & 32.84 & 0.915 & 0.070 & 20.21 \\
\bottomrule
\end{tabular}
}
\end{minipage}
\hfill
\begin{minipage}[t]{0.43\linewidth}
\centering
\caption{Effect of the calibration set size on FFHQ, with the compute time per timestep.}
\label{tab:calibration_images}
\resizebox{0.8\linewidth}{!}{%
\begin{tabular}{c cc cc c}
\toprule
& \multicolumn{2}{c}{Random inp.} & \multicolumn{2}{c}{Box inp.} & \\
\cmidrule(lr){2-3} \cmidrule(lr){4-5}
$N$ & LPIPS & FID & LPIPS & FID & Time (s) \\
\midrule
1   & 0.069 & 19.03 & 0.107 & 44.13 & 0.11 \\
2   & 0.069 & 19.14 & 0.106 & 43.84 & 0.18 \\
5   & 0.069 & 19.18 & 0.106 & 43.61 & 0.38 \\
10  & 0.069 & 19.18 & 0.106 & 43.79 & 0.68 \\
20  & 0.069 & 19.20 & 0.106 & 43.62 & 1.35 \\
50  & 0.069 & 19.17 & 0.106 & 43.71 & 3.37 \\
100 & 0.069 & 19.16 & 0.106 & 43.68 & 6.76 \\
\bottomrule
\end{tabular}
}
\end{minipage}

\end{table*}

\paragraph{Sampler and discretization.}
Recalibrating SIMCA for four samplers across seven step counts (Appendix~\ref{app:samplers_ablation}) shows two patterns. Stochastic samplers dominate the deterministic ones at every budget.
%above $50$ steps, by roughly $1.3$\,dB and $0.025$ LPIPS on random inpainting at $1000$ steps, which reproduces the ordering observed between the renoising and Euler samplers on the flow-matching side. 
They also degrade differently: DDPM improves in PSNR down to $50$ steps and only fails below $20$, whereas Euler-Maruyama collapses below $100$ steps, losing $14$\,dB by $20$. Each cell is calibrated at its own discretization, which is what makes the low-step regime accessible.

\section{Conclusion}
\label{sec:conclusion}
In this work, we introduced SIMCA, a simple and principled offline strategy for tuning guidance weights in training-free posterior sampling. By exploiting the least-squares structure of conditional score- and flow-matching objectives, our method reduces guidance optimization to a two-dimensional linear least-squares problem at each sampling step. The resulting weights are optimized once for a given inverse problem setting, without retraining or fine-tuning the underlying generative model. Evaluated across four datasets and multiple inverse problems, using both diffusion- and flow-matching methods, our approach consistently matches or outperforms state-of-the-art methods.

%We have shown that the guidance weights of training-free posterior samplers need not be tuned by hand. Writing the conditional prediction as a weighted sum of the pretrained field and a measurement-consistency term turns the conditional score-matching and flow-matching objectives into a two-dimensional linear least-squares problem at each step, solved in closed form without ever evaluating the intractable conditional field. The construction is agnostic to the backbone, and SIMCA calibrates under the distribution inference actually induces, at the cost of a single minibatch of trajectories.

%Instantiated with the plain Tweedie-based Gaussian likelihood of DPS, SIMCA %calibration
%matches or surpasses substantially more elaborate samplers across five inverse problems, four datasets, and both backbones, and keeps the sampler effective down to 50 steps. Most tellingly, three different measurement-consistency terms converge to within $0.07$\,dB of one another once calibrated, from a $2.8$\,dB spread: the differences that motivate much of the recent literature are largely differences in effective scaling rather than in expressiveness.

%Where that two-term family is itself inadequate, as on ImageNet Gaussian deblurring, no choice of weights repairs it. Enlarging it, through matrix-valued weighting or a small basis of measurement terms combined by the same argument, would preserve the closed-form solve while spanning directions the present construction cannot reach.

\section*{Use of AI tools}

In this work, we used a large language model to aid and polish writing. Its use was limited to rephrasing and condensing author-written text, drafting figure and table captions, and assisting with LaTeX formatting. We did not use generative AI tools for research ideation, for deriving or verifying the theoretical results, or for writing the code used in our experiments.

\bibliography{iclr2027_conference}
\bibliographystyle{iclr2027_conference}

\appendix
\section*{Appendix}

\section{Extended Related Work}

\paragraph{Control formulation.} \citet{li2024solving} derive a controller that recovers the idealized posterior sampling equation as a special case. Such methods optimize a control signal of full dimension, which requires differentiating through the sampling dynamics at substantial cost. This places them outside the regime we target, where a practitioner needs the guidance to be set in seconds for each new measurement setup.

\paragraph{Exposure bias.}
SIMCA addresses the mismatch between the forward marginals and the states visited at inference, known as exposure bias in diffusion models~\citep{ning2023input, ning2024elucidating}. Epsilon scaling~\citep{ning2024elucidating} rescales the network output with a hand-tuned schedule, an unconditional analogue of our prior weight $a_t$, which we instead obtain from an explicit criterion.

\section{Calibrated weights}
\label{app:weights}

\begin{figure}[H]
\centering
\begin{tikzpicture}
\begin{groupplot}[
    group style={
        group size=2 by 2,
        horizontal sep=1.8cm,
        vertical sep=1.5cm,
    },
    width=7cm, height=6cm,
    xlabel={$t$},
    xticklabels={0.0, 0.0, 0.2, 0.4, 0.6, 0.8, 1.0},
    grid=major,
]

% ---- row 1: FFHQ ----
\nextgroupplot[title={FFHQ: prior weights $w_{u,t}$}, ylabel={$w_{u,t}$}]
\addplot[thick, blue]            tabe[x=t, y=a_p, col sep=comma] {\ffhqfile{box_inpainting}};
\addplot[thick, red]             table[x=t, y=a_p, col sep=comma] {\ffhqfile{random_inpainting}};
\addplot[thick, green!60!black]  table[x=t, y=a_p, col sep=comma] {\ffhqfile{super_resolution}};
\addplot[thick, orange]          table[x=t, y=a_p, col sep=comma] {\ffhqfile{motion_blur}};
\addplot[thick, violet]          table[x=t, y=a_p, col sep=comma] {\ffhqfile{gaussian_blur}};

\nextgroupplot[title={FFHQ: likelihood weights $w_{f,t}$}, ylabel={$w_{f,t}$}, ymode=log]
\addplot[thick, blue]            table[x=t, y=a_l, col sep=comma] {\ffhqfile{box_inpainting}};
\addplot[thick, red]             table[x=t, y=a_l, col sep=comma] {\ffhqfile{random_inpainting}};
\addplot[thick, green!60!black]  table[x=t, y=a_l, col sep=comma] {\ffhqfile{super_resolution}};
\addplot[thick, orange]          table[x=t, y=a_l, col sep=comma] {\ffhqfile{motion_blur}};
\addplot[thick, violet]          table[x=t, y=a_l, col sep=comma] {\ffhqfile{gaussian_blur}};

% ---- row 2: ImageNet ----
\nextgroupplot[title={ImageNet: prior weights $w_{u,t}$}, ylabel={$w_{u,t}$}]
\addplot[thick, blue]            table[x=t, y=a_p, col sep=comma] {\imagenetfile{box_inpainting}};
\addplot[thick, red]             table[x=t, y=a_p, col sep=comma] {\imagenetfile{random_inpainting}};
\addplot[thick, green!60!black]  table[x=t, y=a_p, col sep=comma] {\imagenetfile{super_resolution}};
\addplot[thick, orange]          table[x=t, y=a_p, col sep=comma] {\imagenetfile{motion_blur}};
\addplot[thick, violet]          table[x=t, y=a_p, col sep=comma] {\imagenetfile{gaussian_blur}};

\nextgroupplot[title={ImageNet: likelihood weights $w_{f,t}$}, ylabel={$w_{f,t}$}, ymode=log]
\addplot[thick, blue]            table[x=t, y=a_l, col sep=comma] {\imagenetfile{box_inpainting}};
\addplot[thick, red]             table[x=t, y=a_l, col sep=comma] {\imagenetfile{random_inpainting}};
\addplot[thick, green!60!black]  table[x=t, y=a_l, col sep=comma] {\imagenetfile{super_resolution}};
\addplot[thick, orange]          table[x=t, y=a_l, col sep=comma] {\imagenetfile{motion_blur}};
\addplot[thick, violet]          table[x=t, y=a_l, col sep=comma] {\imagenetfile{gaussian_blur}};

\end{groupplot}
\end{tikzpicture}
\caption{%
  SIMCA calibrated prior weights $w_{u,t}$ (left column) and likelihood weights $w_{f,t}$ (right column, log scale)
  over diffusion step $t$, for FFHQ (top row) and ImageNet (bottom row).
  Colors indicate task:
  \textcolor{blue}{Box inpainting},
  \textcolor{red}{Random inpainting},
  \textcolor{green!60!black}{Super resolution},
  \textcolor{orange}{Motion blur}, and
  \textcolor{violet}{Gaussian blur}.
}
\label{fig:calibration-weights-diff}
\end{figure}
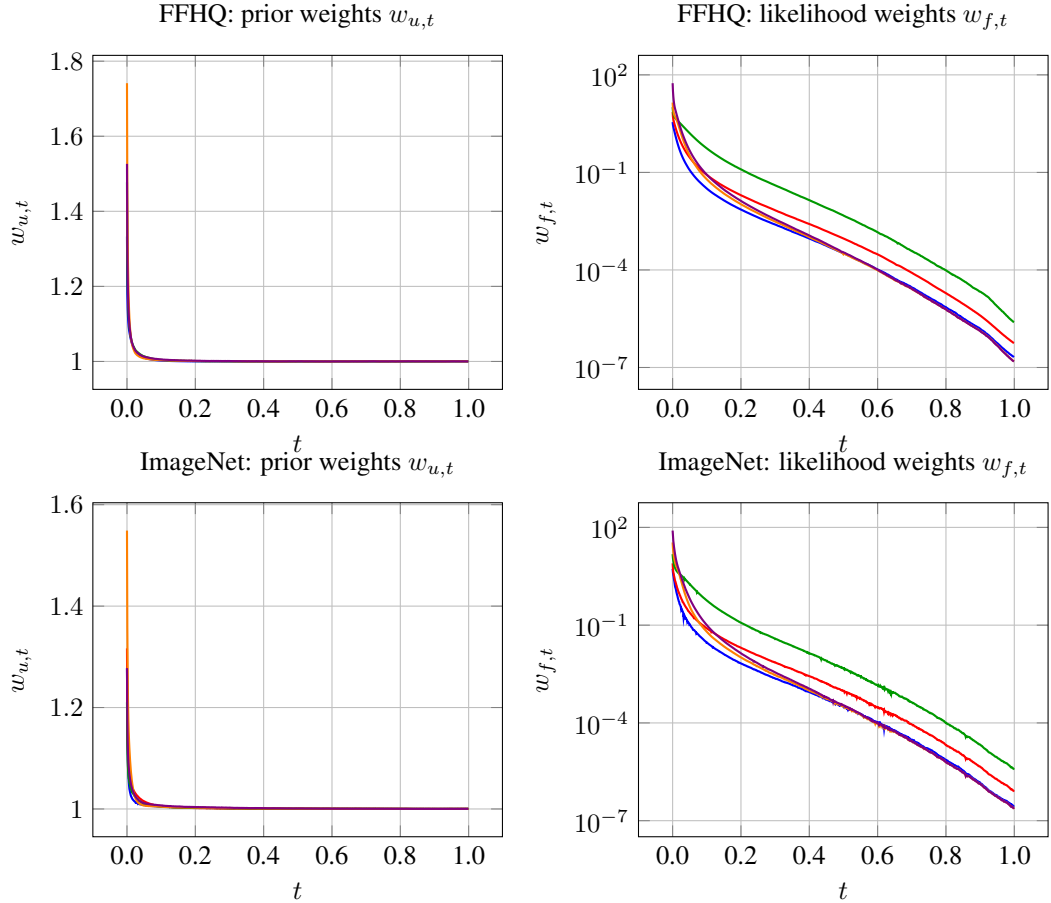

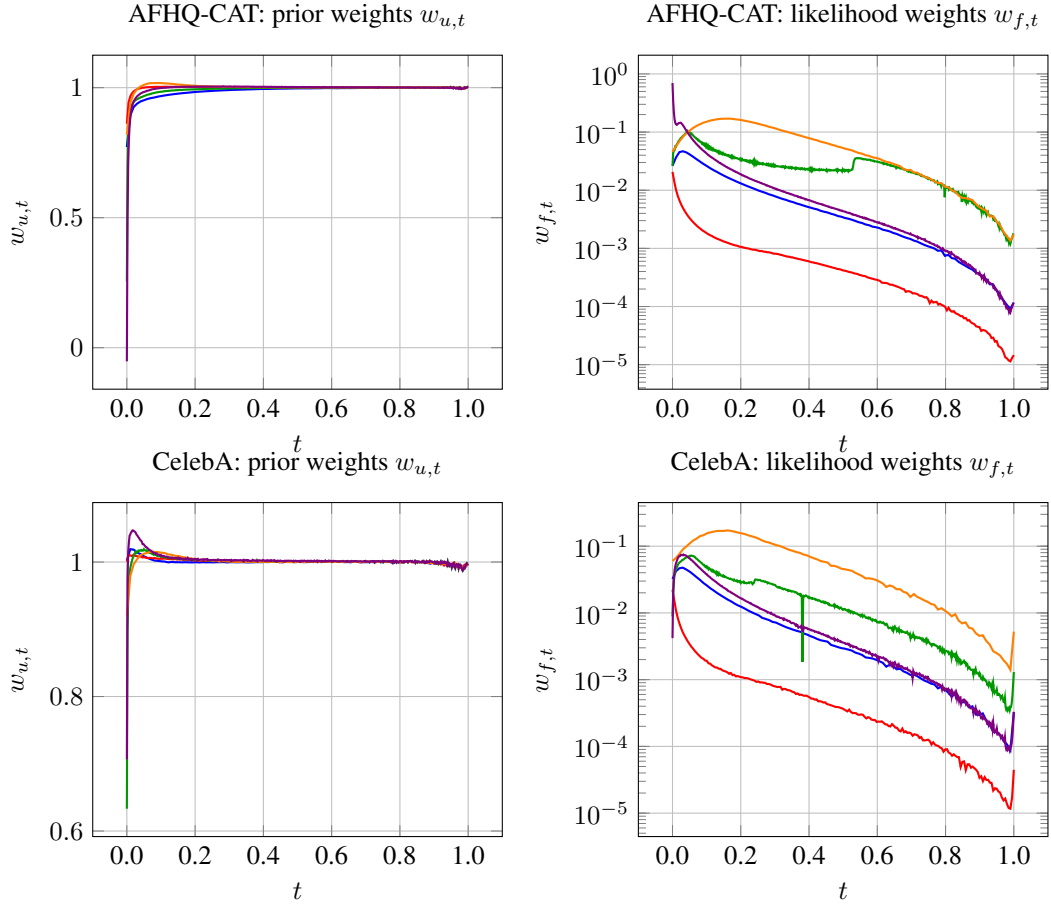
\begin{figure}[H]
\centering
\begin{tikzpicture}
\begin{groupplot}[
    group style={
        group size=2 by 2,
        horizontal sep=1.8cm,
        vertical sep=1.5cm,
    },
    width=7cm, height=6cm,
    xlabel={$t$},
    xticklabels={0.0, 0.0, 0.2, 0.4, 0.6, 0.8, 1.0},
    grid=major,
]

% ---- row 1: AFHQ-CAT ----
\nextgroupplot[title={AFHQ-CAT: prior weights $w_{u,t}$}, ylabel={$w_{u,t}$}]
\addplot[thick, blue]            table[x=t, y=a_v, col sep=comma] {\afhqfile{box_inpainting}};
\addplot[thick, red]             table[x=t, y=a_v, col sep=comma] {\afhqfile{random_inpainting}};
\addplot[thick, green!60!black]  table[x=t, y=a_v, col sep=comma] {\afhqfile{super_resolution}};
\addplot[thick, orange]          table[x=t, y=a_v, col sep=comma] {\afhqfile{denoising}};
\addplot[thick, violet]          table[x=t, y=a_v, col sep=comma] {\afhqfile{gaussian_deblurring}};

\nextgroupplot[title={AFHQ-CAT: likelihood weights $w_{f,t}$}, ylabel={$w_{f,t}$}, ymode=log]
\addplot[thick, blue]            table[x=t, y=a_l, col sep=comma] {\afhqfile{box_inpainting}};
\addplot[thick, red]             table[x=t, y=a_l, col sep=comma] {\afhqfile{random_inpainting}};
\addplot[thick, green!60!black]  table[x=t, y=a_l, col sep=comma] {\afhqfile{super_resolution}};
\addplot[thick, orange]          table[x=t, y=a_l, col sep=comma] {\afhqfile{denoising}};
\addplot[thick, violet]          table[x=t, y=a_l, col sep=comma] {\afhqfile{gaussian_deblurring}};

% ---- row 2: CelebA ----
\nextgroupplot[title={CelebA: prior weights $w_{u,t}$}, ylabel={$w_{u,t}$}]
\addplot[thick, blue]            table[x=t, y=a_v, col sep=comma] {\celebafile{box_inpainting}};
\addplot[thick, red]             table[x=t, y=a_v, col sep=comma] {\celebafile{random_inpainting}};
\addplot[thick, green!60!black]  table[x=t, y=a_v, col sep=comma] {\celebafile{super_resolution}};
\addplot[thick, orange]          table[x=t, y=a_v, col sep=comma] {\celebafile{denoising}};
\addplot[thick, violet]          table[x=t, y=a_v, col sep=comma] {\celebafile{gaussian_deblurring}};

\nextgroupplot[title={CelebA: likelihood weights $w_{f,t}$}, ylabel={$w_{f,t}$}, ymode=log]
\addplot[thick, blue]            table[x=t, y=a_l, col sep=comma] {\celebafile{box_inpainting}};
\addplot[thick, red]             table[x=t, y=a_l, col sep=comma] {\celebafile{random_inpainting}};
\addplot[thick, green!60!black]  table[x=t, y=a_l, col sep=comma] {\celebafile{super_resolution}};
\addplot[thick, orange]          table[x=t, y=a_l, col sep=comma] {\celebafile{denoising}};
\addplot[thick, violet]          table[x=t, y=a_l, col sep=comma] {\celebafile{gaussian_deblurring}};

\end{groupplot}
\end{tikzpicture}
\caption{%
  SIMCA calibrated prior weights $w_{u,t}$ (left column) and likelihood weights $w_{f,t}$ (right column, log scale)
  over diffusion step $t$, for AFHQ-CAT (top row) and CelebA (bottom row).
  Colors indicate task:
  \textcolor{blue}{Box inpainting},
  \textcolor{red}{Random inpainting},
  \textcolor{green!60!black}{Super resolution},
  \textcolor{orange}{Denoising}, and
  \textcolor{violet}{Gaussian blur}.
}
\label{fig:calibration-weights-flow}
\end{figure}

Figures~\ref{fig:calibration-weights-diff} and~\ref{fig:calibration-weights-flow} report the calibrated weights $(w_{u,t}, w_{f,t})$ for additional datasets and operators, on the diffusion and flow-matching backbones respectively.

% \documentclass{article}

% \usepackage{pgfplots}
% \usepackage{xcolor}
% \pgfplotsset{compat=1.18}
% \usepgfplotslibrary{groupplots}

% \newcommand{\ffhqfile}[1]{a_star_weights_1000_100_#1.csv}
% \newcommand{\imagenetfile}[1]{a_star_weights_1000_50_imagenet_#1.csv}

%\begin{document}

%\end{document}

%\end{document}

\section{Effect of the step count}
\label{app:step_count}

\begin{figure}[h]
    \centering
    \includegraphics[width=\textwidth]{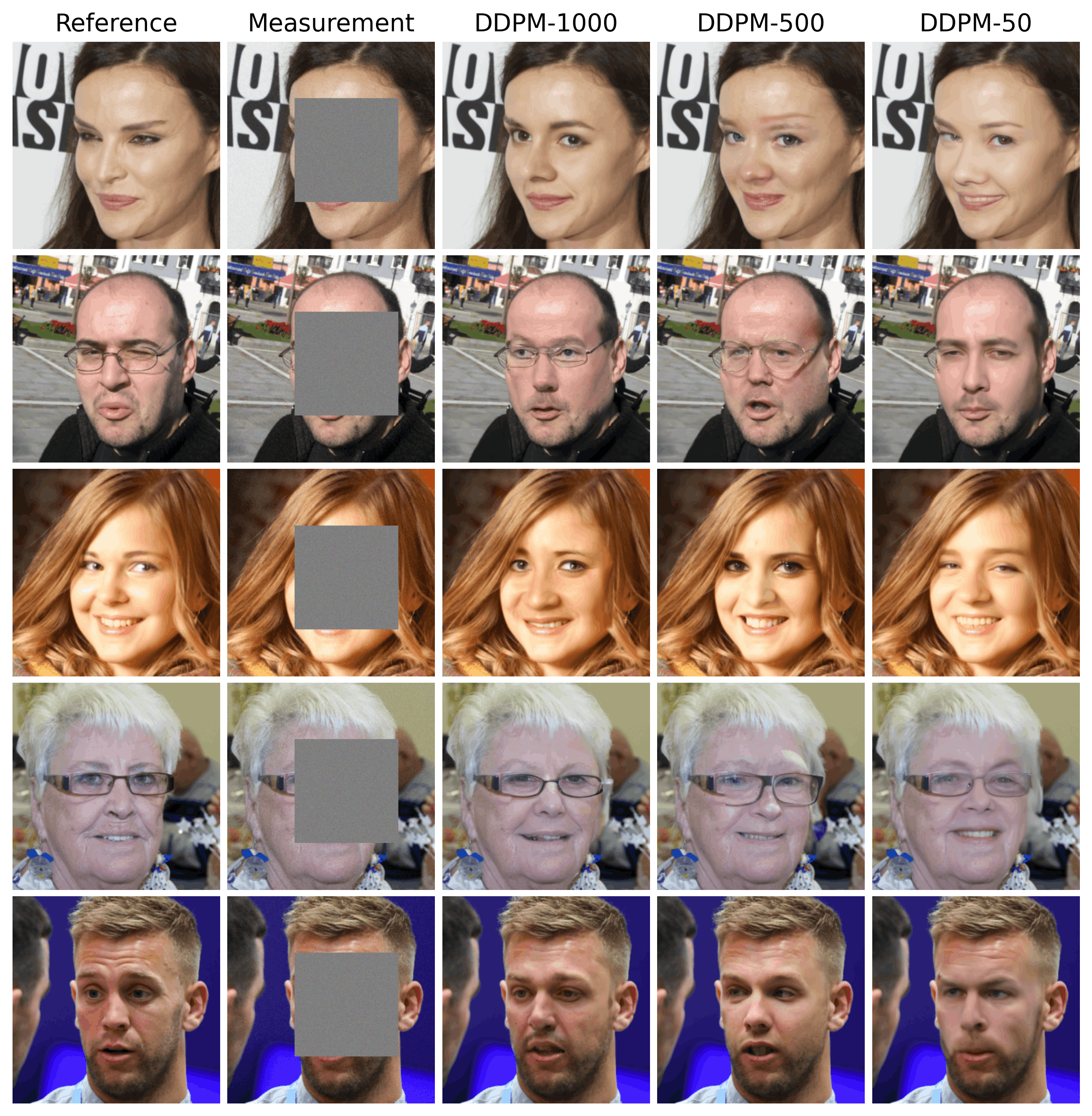}
    \caption{SIMCA on FFHQ box inpainting at $K = 1000$, $500$ and $50$ steps, recalibrated at each discretization.}
    \label{fig:step_count_bi}
\end{figure}

\begin{figure}[h]
    \centering
    \includegraphics[width=\textwidth]{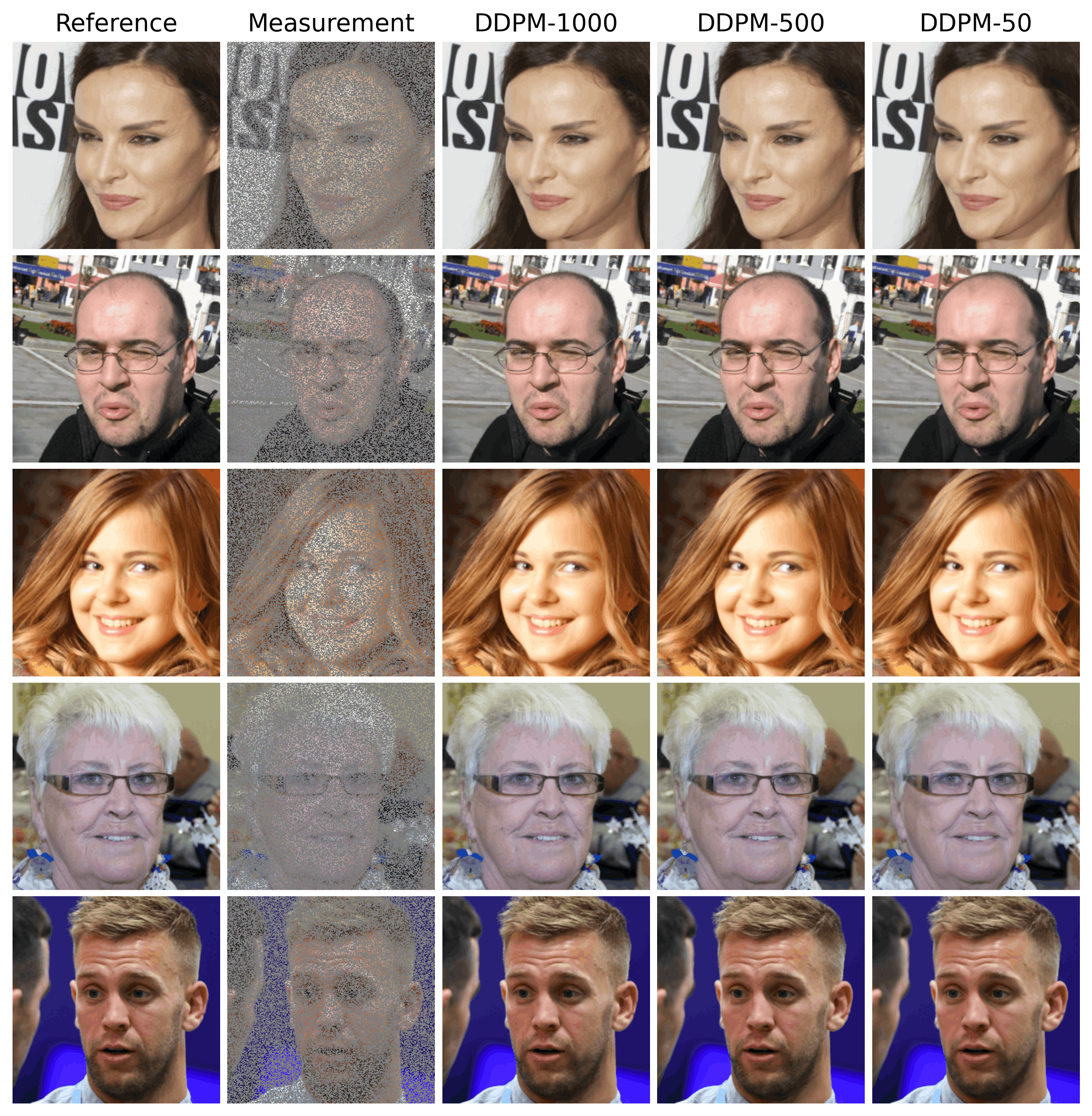}
    \caption{SIMCA on FFHQ random inpainting at $K = 1000$, $500$ and $50$ steps, recalibrated at each discretization.}
    \label{fig:step_count_ri}
\end{figure}

\begin{figure}[h]
    \centering
    \includegraphics[width=\textwidth]{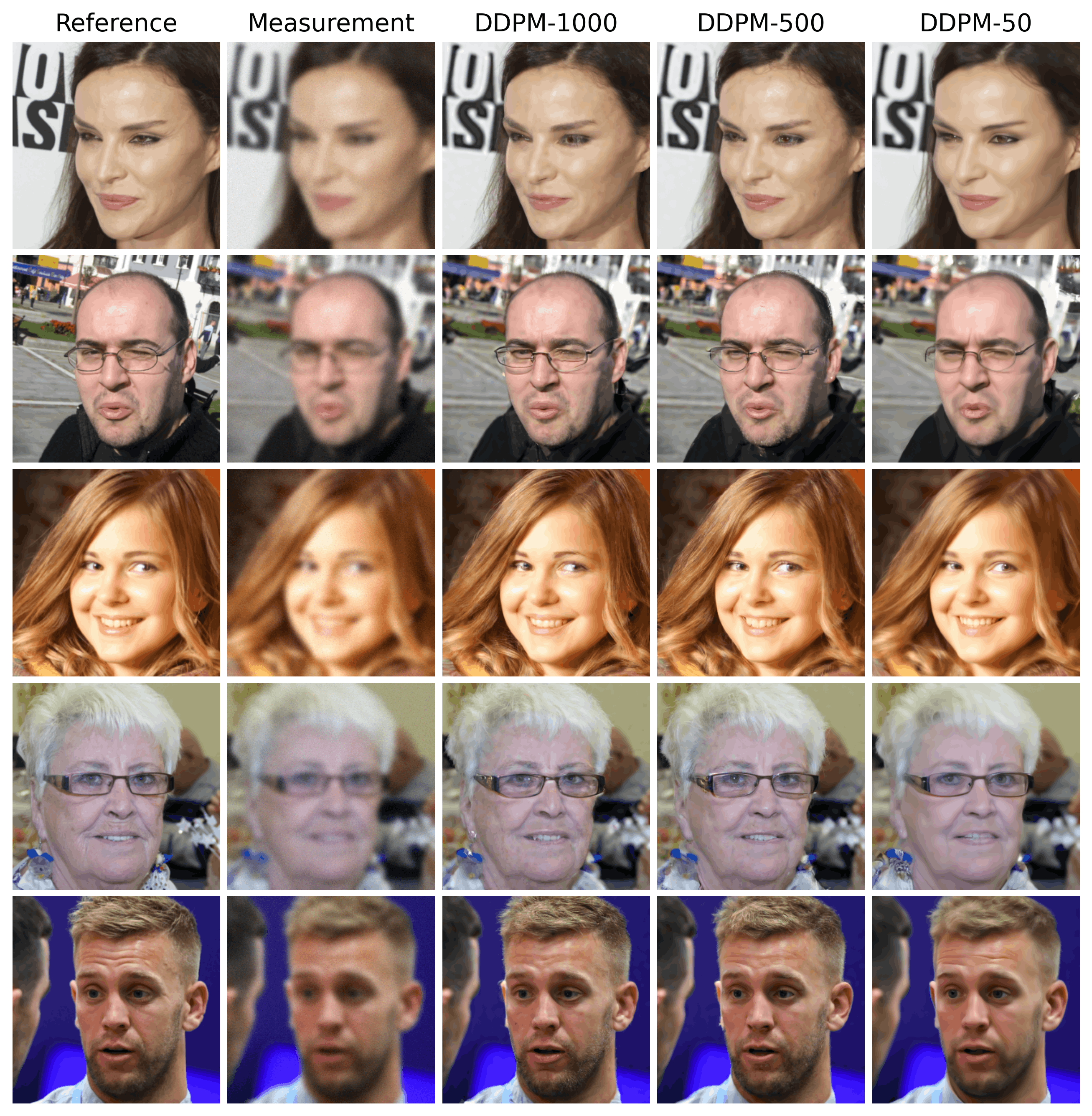}
    \caption{SIMCA on FFHQ Gaussian deblurring at $K = 1000$, $500$ and $50$ steps, recalibrated at each discretization.}
    \label{fig:step_count_gb}
\end{figure}

\begin{figure}[h]
    \centering
    \includegraphics[width=\textwidth]{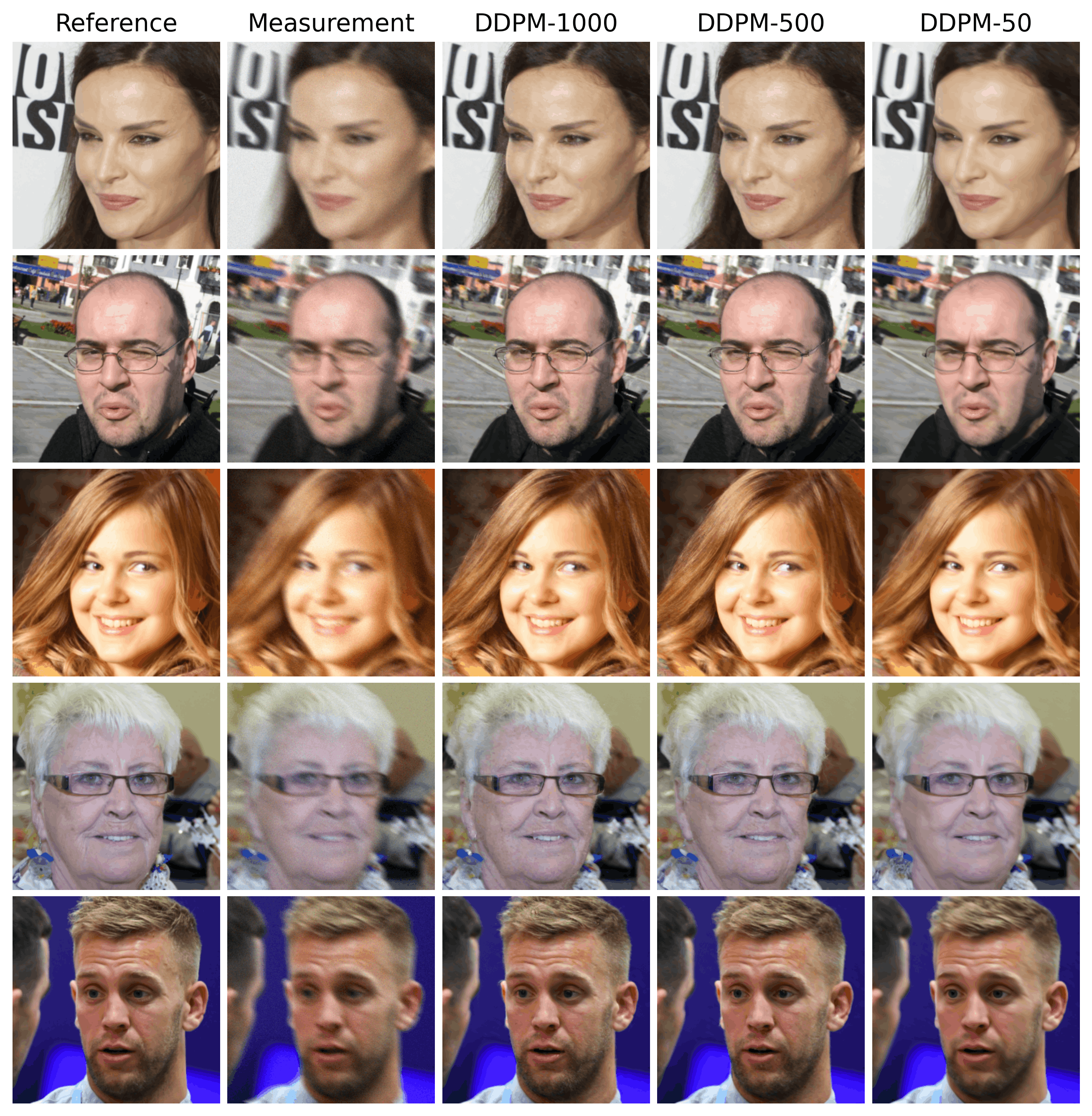}
    \caption{SIMCA on FFHQ motion deblurring at $K = 1000$, $500$ and $50$ steps, recalibrated at each discretization.}
    \label{fig:step_count_mb}
\end{figure}

\begin{figure}[h]
    \centering
    \includegraphics[width=\textwidth]{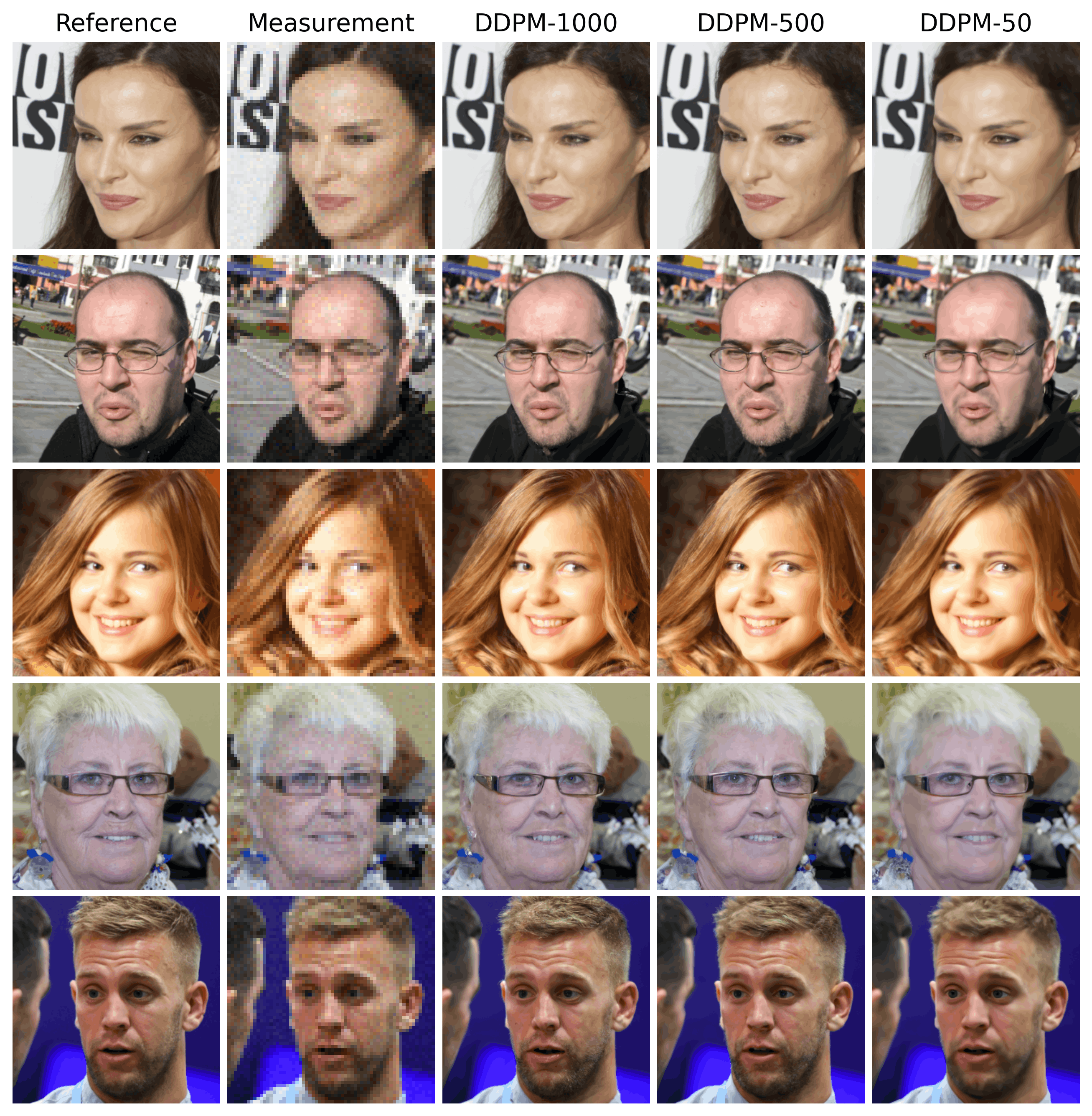}
    \caption{SIMCA on FFHQ super-resolution at $K = 1000$, $500$ and $50$ steps, recalibrated at each discretization.}
    \label{fig:step_count_sr}
\end{figure}

Figures~\ref{fig:step_count_bi} to~\ref{fig:step_count_sr} show SIMCA reconstructions at $K = 1000$, $500$ and $50$ steps, recalibrated at each discretization, on the five diffusion tasks. On the four well-posed operators (Gaussian deblurring, motion deblurring, super-resolution, random inpainting) the three columns are visually almost indistinguishable, which is the qualitative counterpart of the metric behaviour reported in Section~\ref{sec:quant}: reducing the budget twentyfold leaves reconstruction quality essentially unchanged. Box inpainting behaves differently, and instructively so. There the three budgets produce visibly different completions of the masked region, since that region is not determined by the measurement, while the unmasked part remains identical across columns. The step count therefore acts as a choice among plausible posterior samples rather than as a loss of fidelity, and the $50$-step reconstructions are no less plausible than the $1000$-step ones, consistent with their higher PSNR and slightly worse FID.

\section{Robustness to the backbone and the sampler.}
\label{app:robust}

In Tab.~\ref{tab:celeba} and Tab.~\ref{tab:afhq_cat}, two comparisons isolate what the gains depend on. First, in the flow-matching experiments, the OT-coupled and independently coupled backbones behave almost identically once calibrated, differing by at most $0.2$\,dB and $0.002$ LPIPS on CelebA. %Therefore, SIMCA 
%calibration
%does not rely on a particular pretrained field, and in particular not on the OT-coupling the strongest baselines are built on. 
Second, the renoising sampler outperforms the plain Euler integrator by $3.7$\,dB on CelebA random inpainting and $3.0$\,dB on box inpainting. This matches the deterministic-sampler penalty of Section~\ref{sec:ablations} and is a property of the sampler rather than of calibration: the Euler rows still improve on most flow matching baselines.

\section{Sampler and discretization}
\label{app:samplers_ablation}

Table~\ref{tab:ablation_samplers} reports SIMCA recalibrated for four samplers across seven step counts on FFHQ. The two families degrade very differently as the budget shrinks: DDPM improves in PSNR down to $50$ steps and only fails below $20$, whereas Euler-Maruyama collapses below $100$ steps, since its explicit discretization of the reverse SDE is unstable at large step sizes. DDIM degrades gracefully, remaining usable at $10$ steps, at a consistently lower level throughout. The practical recommendation is DDPM with recalibrated weights at $50$ steps, which gives the best distortion metrics of the table on random inpainting and super-resolution at a twentieth of the nominal cost. The table also quantifies the dependence noted in Section~\ref{sec:method}: since the weights depend on the sampler $\Phi$ and on the number of timesteps, transferring them across these rows rather than recalibrating would forfeit the gains.

\begin{table*}[!]
    \caption{Effect of the sampler and the discretization, on FFHQ. SIMCA is recalibrated independently for each sampler and step count $K$. Stochastic samplers (DDPM, Euler-Maruyama) outperform their deterministic counterparts (DDIM, Euler-ODE) at every budget, and DDPM alone remains stable down to $K=50$ where it attains the best distortion metrics of the table on random inpainting and super-resolution.}
  \label{tab:ablation_samplers}
  \centering
  \setlength{\tabcolsep}{2pt}
  \resizebox{\textwidth}{!}{%
  \small
  \begin{tabular}{c|c|cccc|cccc|cccc}
    \toprule
     & & \multicolumn{4}{c}{Random Inpainting} & \multicolumn{4}{c}{Box Inpainting} & \multicolumn{4}{c}{SRx4} \\
    Samplers & K & PSNR & SSIM & LPIPS & FID & PSNR & SSIM & LPIPS & FID & PSNR & SSIM & LPIPS & FID \\ \midrule
    \multirow{7}{*}{DDPM} 
    & 1000 & 
    32.91 & 0.916 & 0.069 & 19.16 & 
    22.90 & 0.846 & 0.106 & 43.68 & 
    29.12 & 0.821 & 0.145 & 46.39 \\
    & 500 & 
    33.07 & 0.920 & 0.068 & 18.31 & 
    23.28 & 0.850 & 0.105 & 39.77 & 
    29.26 & 0.826 & 0.143 & 43.14  \\
    & 200 & 
    33.34 & 0.926 & 0.068 & 20.76 & 
    23.28 & 0.853 & 0.106 & 43.26 & 
    29.48 & 0.836 & 0.141 & 44.51 \\
    & 100 & 
    33.62 & 0.930 & 0.068 & 23.04 
    & 23.36 & 0.856 & 0.108 & 44.52 & 
    29.71 & 0.845 & 0.140 & 45.50 \\
    & 50 & 
    33.76 & 0.930 & 0.072 & 24.88 & 
    23.43 & 0.855 & 0.110 & 47.12 & 
    30.00 & 0.853 & 0.144 & 49.02 \\
    & 20 & 
    33.11 & 0.918 & 0.090 & 33.60 & 
    22.94 & 0.847 & 0.122 & 51.02 & 
    30.12 & 0.855 & 0.156 & 53.89 \\
    & 10 & 
    31.06 & 0.886 & 0.132 & 59.91 & 
    23.13 & 0.841 & 0.140 & 54.67 & 
    29.50 & 0.833 & 0.195 & 74.42 \\
    \midrule
    \multirow{7}{*}{Euler-SDE} 
    & 1000 & 
    32.88 & 0.915 & 0.069 & 18.99 & 
    22.90 & 0.845 & 0.106 & 44.07 
    & 29.11 & 0.820 & 0.146 & 46.76  \\
    & 500 & 
    33.04 & 0.919 & 0.068 & 18.41 & 
    23.25 & 0.848 & 0.105 & 40.67 & 
    29.26 & 0.825 & 0.144 & 43.60  \\
    & 200 & 
    32.97 & 0.914 & 0.075 & 23.14 & 
    23.12 & 0.837 & 0.114 & 46.17 & 
    29.42 & 0.828 & 0.145 & 43.01  \\
    & 100 & 
    32.20 & 0.883 & 0.109 & 31.91 & 
    22.88 & 0.808 & 0.143 & 57.17 & 
    29.34 & 0.809 & 0.160 & 45.54  \\
    & 50 & 
    30.03 & 0.780 & 0.198 & 51.64 & 
    22.11 & 0.731 & 0.201 & 84.24 & 
    28.44 & 0.723 & 0.220 & 53.39  \\
    & 20 & 
    24.57 & 0.462 & 0.404 & 96.08 & 
    19.69 & 0.535 & 0.352 & 186.90 & 
    24.77 & 0.462 & 0.382 & 87.39  \\
    & 10 & 
    18.43 & 0.192 & 0.618 & 184.61 & 
    17.46 & 0.315 & 0.505 & 270.25 & 
    20.79 & 0.262 & 0.512 & 122.34  \\
    \midrule
    \multirow{7}{*}{DDIM} 
    & 1000 & 
    31.56 & 0.901 & 0.094 & 32.02 & 
    21.58 & 0.825 & 0.136 & 63.93 & 
    28.50 & 0.797 & 0.176 & 58.31  \\
    & 500 & 
    32.04 & 0.908 & 0.086 & 28.35 & 
    21.97 & 0.832 & 0.125 & 55.03 & 
    28.65 & 0.796 & 0.174 & 56.06 \\
    & 200 & 
    32.24 & 0.913 & 0.084 & 28.26 & 
    21.95 & 0.833 & 0.125 & 55.07 & 
    28.70 & 0.794 & 0.177 & 55.97 \\
    & 100 & 
    32.49 & 0.917 & 0.083 & 28.54 & 
    21.95 & 0.833 & 0.126 & 55.11 & 
    28.81 & 0.795 & 0.179 & 55.87 \\
    & 50 & 
    32.69 & 0.918 & 0.084 & 29.88 & 
    21.97 & 0.832 & 0.128 & 55.07 & 
    28.86 & 0.788 & 0.185 & 55.22 \\
    & 20 & 
    32.08 & 0.905 & 0.106 & 41.88 & 
    21.99 & 0.828 & 0.139 & 57.09 & 
    28.44 & 0.741 & 0.224 & 61.46 \\
    & 10 & 
    29.99 & 0.866 & 0.157 & 69.46 & 
    21.90 & 0.820 & 0.156 & 62.81 & 
    27.48 & 0.674 & 0.279 & 80.06 \\
    \midrule
    \multirow{7}{*}{Euler-ODE} 
    & 1000 & 
    31.77 & 0.902 & 0.090 & 29.32 & 
    21.82 & 0.827 & 0.127 & 56.81 & 
    28.58 & 0.787 & 0.187 & 57.18 \\
    & 500 & 
    31.66 & 0.900 & 0.092 & 30.22 & 
    21.83 & 0.825 & 0.126 & 58.04 & 
    28.48 & 0.779 & 0.188 & 56.15 \\
    & 200 & 
    31.27 & 0.889 & 0.101 & 34.71 & 
    21.81 & 0.814 & 0.138 & 64.12 & 
    28.27 & 0.760 & 0.193 & 54.63 \\
    & 100 & 
    30.38 & 0.856 & 0.136 & 44.44 & 
    21.68 & 0.781 & 0.170 & 81.52 & 
    27.98 & 0.733 & 0.204 & 54.73 \\
    & 50 & 
    28.16 & 0.734 & 0.236 & 64.77 & 
    21.13 & 0.694 & 0.252 & 128.57 & 
    27.22 & 0.664 & 0.246 & 60.99 \\
    & 20 & 
    22.16 & 0.359 & 0.479 & 108.41 & 
    21.89 & 0.438 & 0.463 & 240.49 & 
    23.34 & 0.376 & 0.434 & 95.71 \\
    & 10 & 
    15.92 & 0.119 & 0.684 & 213.88 & 
    15.73 & 0.189 & 0.632 & 280.34 & 
    17.90 & 0.150 & 0.616 & 150.32 \\

  \end{tabular}
  \small
  }
\end{table*}

\section{Experiments - technical details}\label{app:exp_details}
\paragraph{Backbones and protocols.}
For diffusion, we use the pixel-space DDPM checkpoints for FFHQ and ImageNet at $256\times256$ released by~\citep{chung2022diffusion}, following the evaluation protocol of DAPS~\citep{zhang2025improving}. For flow matching, we adopt the setup of Flower~\citep{pourya2026flower}, using the same U-Net velocity networks trained on CelebA and AFHQ-Cat. We evaluate both the OT-coupled network and its independently coupled variant, allowing us to assess calibration across two distinct pretrained velocity fields.

\paragraph{Datasets and metrics.}
We evaluate on 100 held-out images per dataset: FFHQ and ImageNet at $256\times256$ for diffusion, CelebA at $128\times128$ and AFHQ-Cat at $256\times256$ for flow matching. Images are scaled to $[-1,1]$ for sampling and noise levels are expressed on this scale. We report PSNR, SSIM, and LPIPS~\citep{zhang2018unreasonable} on both benchmarks, and FID~\citep{heusel2017gans} on the diffusion benchmark, averaged over individual images.

\paragraph{Inverse problems.}
On the diffusion side we consider super-resolution ($\times4$ bicubic downsampling), Gaussian deblurring ($61\times61$ kernel, intensity $3.0$), motion deblurring ($61\times61$ kernel, intensity $0.5$), box inpainting ($128\times128$ central mask), and random inpainting ($70\%$ masking rate), with additive white Gaussian noise of standard deviation $\sigma_m = 0.05$. On the flow-matching side we consider the five tasks of the Flower benchmark, with operators and noise levels unchanged: denoising, Gaussian deblurring, super-resolution, random inpainting, and box inpainting.

\end{document}